\documentclass[times,twocolumn,final]{elsarticle} 
\usepackage{prletters}
\usepackage{balance}
\usepackage{amssymb}
\usepackage{latexsym}

\usepackage{url}
\usepackage{xcolor}
\definecolor{newcolor}{rgb}{.8,.349,.1}

\usepackage[]{algorithm2e}

\usepackage{caption}
\usepackage{subcaption}
\graphicspath{{/images/}}
\usepackage[retainorgcmds]{IEEEtrantools}
\usepackage{mdframed}
\usepackage[english]{babel}

\usepackage{array}
\newcolumntype{C}[1]{>{\centering\let\newline\\\arraybackslash\hspace{0pt}}m{#1}}

\begin{document}

%
%

\begin{abstract}
The problem of detecting changes in a scene and segmenting the foreground from background is still challenging, despite previous work. 
Moreover, new RGBD capturing devices include depth cues, which could be incorporated to improve foreground segmentation. In this work, we present a new nonparametric approach where a unified model mixes the device multiple information cues. In order to unify all the device channel cues, a new probabilistic depth data model is also proposed where we show how handle the inaccurate data to improve foreground segmentation. 
A new RGBD video dataset is presented in order to introduce a new standard for comparison purposes of this kind of algorithms. Results show that the proposed approach can handle several practical situations and obtain good results in all cases.

\end{abstract}

\begin{frontmatter}

\title{Modelling depth for nonparametric foreground segmentation using RGBD devices}

\author{Gabriel {Moy\`a-Alcover}} 
\author{Ahmed {Elgammal}}
\author{Antoni {Jaume-i-Cap\'o}}
\author{Javier {Varona}}

\end{frontmatter}



\section{Introduction}

Background subtraction is a widely used technique for detecting moving foreground objects in image sequences. It is 
considered the first step in many computer vision algorithms. Foreground segmentation, provides an important cue for numerous applications in computer vision, such as surveillance, tracking, recognition and human pose estimation. The main objective is to detect objects that do not belong to the scene by comparing the current observation with previous references. This reference can be a single image or a more complex model of the real scene, called \textit{scene model}. A scene model is a statistical representation of the scene, and it is updated to adapt to variations of its conditions.

This problem has been widely addressed in the literature. Reviews can be found in~\cite{bouwmans2014background, elgammal2014background}. 
Despite this previous research, there is no universal technique covering all requirements of applications for which the foreground of a scene must be detected~\cite{Bouwmans2014}. In~\cite{Toyama1999} several important challenges of background subtraction were described. Some of them are strongly related to the nature of color information, such as: shadows, changes in scene illumination, camouflage and foreground aperture. 
These problems continue to be challenging for modern approaches, as described in~\cite{Sobral2014}, where 29 different algorithms were evaluated and compared.

A feasible solution to overcome these problems consists of adding physical information to the background model. For example, geometrical descriptions of buildings may be added to help to predict shadows~\cite{rogez2013prior}.     

Different approaches for obtaining 3D information of the scene were proposed using stereo devices or camera networks~\cite{Cristani2010}. 
Depth measures provide geometrical information about the scene where each pixel value represents the distance from the device to the point in the real world. To obtain an accurate dense map of correlations between two stereo images, time-consuming stereo algorithms are required. Without specialized hardware, most of these algorithms are too slow for real-time background subtraction. In addition, multi-camera networks introduce other problems, such as camera installation, calibration and data fusion.

Currently, low-cost RGBD devices that are able to capture depth and color images simultaneously at frame rates up to 30 fps are available off the shelf. These devices have certain limitations such as lower sensitivity at long distances, the production of depth camouflage and absent observations due to scene characteristics.

Our aim is to use this type of noisy depth information in a unified model that mixes multiple information cues from the devices. We present a 
new per-pixel scene modeling approach which uses both depth and color information. We propose a model that keeps a sample for each pixel of 
the scene and estimates the probability that a newly observed pixel value belongs to the background. The model estimates these 
probabilities independently for each new frame. The model is updated in each iteration of the algorithm, depending on partial results. The 
model adapts itself to changes in the background process and detects targets with high sensitivity.

We construct our model using a Kernel Density Estimation (KDE) process. KDE has been already used in other 
\textit{state-of-the-art} techniques. In particular, in~\cite{Elgammal2000}, KDE has been applied using only color information with good results. When using a Gaussian Kernel, the probability density function can be thought of as a generalization of the Gaussian mixture model, in which each single sample is considered to be a Gaussian distribution by itself. However unlike in the Gaussian Mixture Model, in KDE no mixture parameters need to be estimated. This allows us to estimate the density function more accurately, without assumptions about the density model, depending only on recent information from the sequence.

Adding a depth channel to the KDE background model is not an obvious process because the depth channel differs in its characteristics from color channels. In particular, the depth channel has a significant amount of missing information from instances in which  the sensor is unable to estimate the depth  at certain pixels. In this paper, we show how to handle the inaccurate depth data in the proposed nonparametric scene model. For this purpose, we properly define the absent depth observations to include them in the scene model. The key idea is that pixels that cannot be classified as background or foreground are classified in a new undefined class. Therefore, absent observations can be handled in a unified manner. In addition, after the introduction of depth data, the proposed scene model is capable of instantly detecting the changes in the background objects.

To properly evaluate the proposed method, we built a new dataset inspired by one of the most widely used color-based datasets~\cite{Toyama1999}. Each of the proposed sequences is focused on one of the main challenges when both color and depth information are used.

The paper is organized as follows. In Section~\ref{sec:StateOfArt}, we describe the related work. In Section~\ref{sec:challengesOfDepthData}, we describe the challenges of depth data. In Section \ref{sec:BackgroundModel}, we define the proposed scene model, and we explain how depth information is used to construct a unified model. Adaptation to scene changes is discussed in Section~\ref{sec:ModelUpdate}. In Section~\ref{sec:Exp}, we describe an experimental configuration of the proposed algorithm. The results of the evaluation are described in Section~\ref{sec:evaluation}. Finally, we present the conclusions.

\section{Related work}
\label{sec:StateOfArt}
There is a large body of literature on the subject of background subtraction. We refer to some comprehensive surveys about this 
subject~\cite{bouwmans2014background,elgammal2014background, Cristani2010, Benezeth2012, Sobral2014}. We focus here on approaches that fuse color and depth information.  Most of these techniques modify traditional background subtraction approaches by adding one extra channel for depth (in addition to the color channels) and suggesting some heuristics to address the heterogeneous characteristics of these different cues.

In~\cite{Harville2001}, the authors proposed an approximation to Gaussian mixture modeling to describe the recent history of color and depth scene observations at each pixel. A multidimensional Gaussian mixture distribution is constructed, with three components in a luminance-normalized color space and one depth channel. Special processing is performed to address absent depth pixels. This enables foreground decisions to be made when the depth model for a pixel is invalid but its latest depth observation is valid and it is connected to regions where foreground decisions have been made in the presence of valid background data. No update phase is described; therefore, this algorithm can only be used in static scenes.

In~\cite{Hofmann}, a new Mixture of Gaussians approach is proposed, where depth and infrared data are combined to detect foreground objects. Two independent background models are built using depth and infrared information. Each pixel is classified by binary combinations of foreground masks. The performance of this approach is limited because a failure of one of the models affects the final pixel classification.

Camplani \textit{et al}.~\cite{Camplani2014} proposed a \textit{per-pixel} background modeling approach that fuses different statistical classifiers based on depth and color data by means of a weighted average combination that takes into account the characteristics of depth and color data. A mixture of Gaussian distributions is used to model the background pixels, and a uniform distribution is used for modeling the foreground. The same authors presented another approach in~\cite{Camplani2014a} based on the fusion of multiple region-based classifiers.  Foreground objects are detected by combining a region-based foreground depth data prediction with different background models, providing color and depth descriptions of the scene at the pixel and region levels. The information given by these modules is fused in a mixture-of-experts fashion to improve the foreground detection accuracy.

ViBe is a \textit{per-pixel} algorithm, based on a Parzen windows-like process~\cite{Barnich2011}. The update is performed by a random process that substitutes old pixel values with new ones and then samples the spatial neighbourhoods to refine the per-pixel estimates. ViBe gives acceptable detection results in many scenarios, but it has problems with challenging scenarios such as darker backgrounds, shadows and frequent background changes~\cite{Bouwmans2014}. In~\cite{JeromeLeensSebastienPierardOlivierBarnich1978}, a new ViBe approach is presented using RGB and ToF (\textit{Time-of-Flight}) cameras. Each model is processed independently and the foreground masks are then combined using logical operations and then post-processed with morphological operators.

An adaptation of the Codebook~\cite{Kim2004} background subtraction algorithm was proposed by~\cite{Fernandez-Sanchez2013} 
fusing depth and color information to segment foreground regions. A four-channel codebook was used. Depth information is also used to bias the distance in chromaticity space associated with a pixel according to the depth measurements. Therefore, when the depth value is invalid, the detection depends entirely on color information. Their results were tested on a public access database with four different challenging sequences. For each frame in the dataset, depth information was normalized from 0 to 255, where 255 is the maximum depth value in that frame, with the resulting loss of information.

In~\cite{Clapes2013} the authors presented a background subtraction technique in which a four-dimensional Gaussian distribution was used as the first step of the user identification and object recognition surveillance system. No special processing was performed to address absent depth observation pixels. As they used a single Gaussian approximation, the algorithm was not able to manage multi-modal backgrounds. A similar problem can be observed in other approaches, such as~\cite{Harville2001} and~\cite{Kolmogorov}.

In the related work there is no general purpose RGBD dataset that covers all of the desirable types of sequences, with which to properly evaluate a scene modeling algorithm. Each algorithm is evaluated using its own dataset and different metrics. That makes it impossible to perform a unified comparison between the different methods. For that purpose, we propose a comprehensive dataset that covers the challenges that occur when combining depth and color information.

\section{Challenges of depth data}
\label{sec:challengesOfDepthData}

Depth sensors provide partial geometrical information about the scene, where each pixel depth value is proportional to the estimated distance from the device to the point in the real world. Among several technologies, recently, two types of consumer depth sensors have become widely popular and accessible: sensors based on structured light and on time-of-flight.

\textbf{Structured light sensors} consists of an infrared (IR) emitter and an IR camera. It estimates depth by structured light coding technology. Its IR emitter projects an IR speckle pattern onto the scene. The IR camera captures the reflected pattern and correlates it against a stored reference pattern on a plane. These sensors have a lack of sensitivity and are not able to estimate depth at all pixels in the scene. The noise in depth measurements increases quadratically with increasing distance from the sensor~\cite{Khoshelham2012}.

\textbf{Time-of-flight sensors} resolve the distance based on the known speed of light. Depth is proportional to the time needed by the active illumination source to travel from emitter to target. Typically, IR light is used for this purpose. This technology provides better accuracy than structured light sensors and is less susceptible to generate shadows in the scene. Noise can be well approximated by means of a normal distribution~\cite{fankhauserkinect}.

\begin{figure}[t]
    \begin{subfigure}[b]{0.5\textwidth}
    		{\includegraphics[height=0.7in]{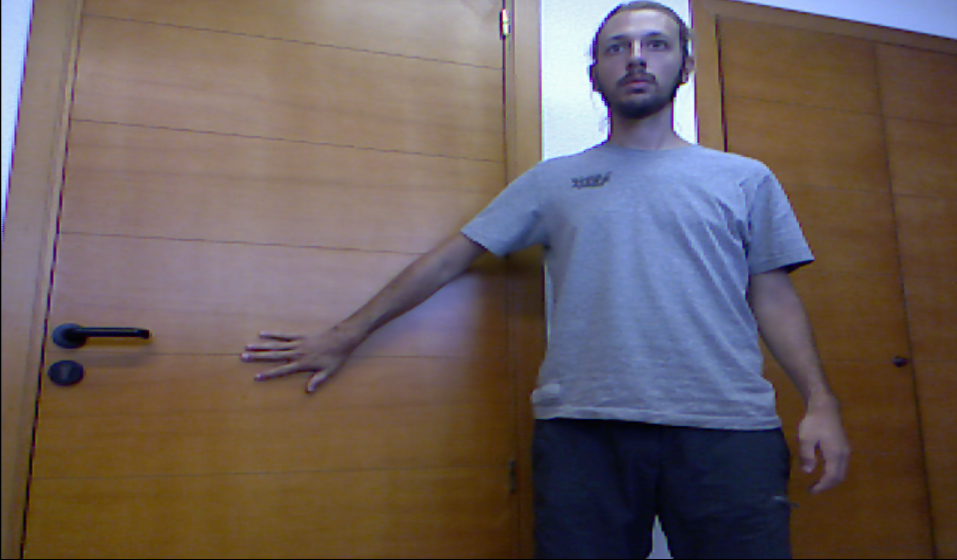}}
   		{\includegraphics[height=0.7in]{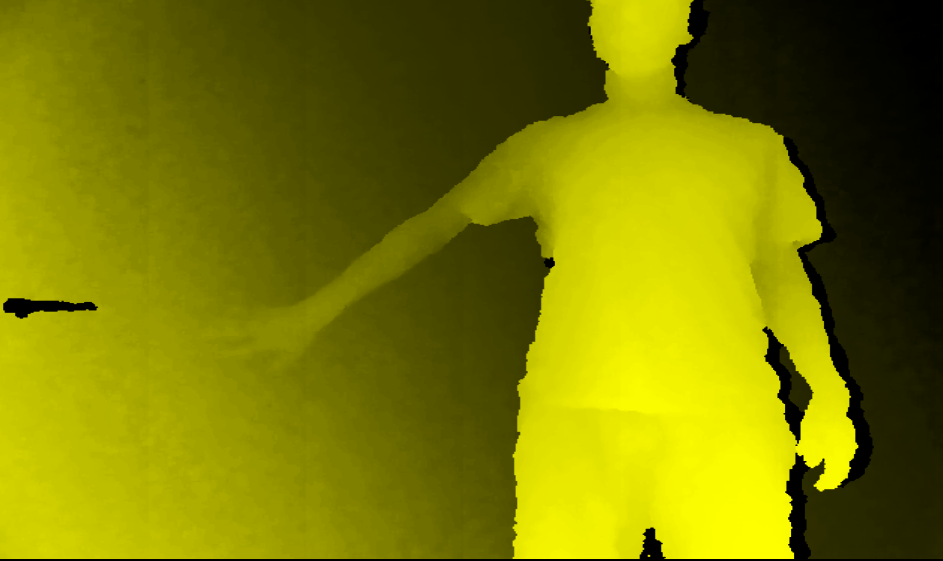}}
        {\includegraphics[height=0.7in]{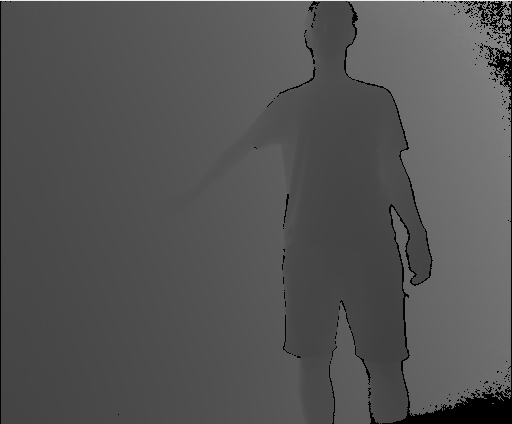}}
         \caption{Depth camouflage.}
    \end{subfigure}\\
     \begin{subfigure}[b]{0.5\textwidth}
         {\includegraphics[height=0.7in]{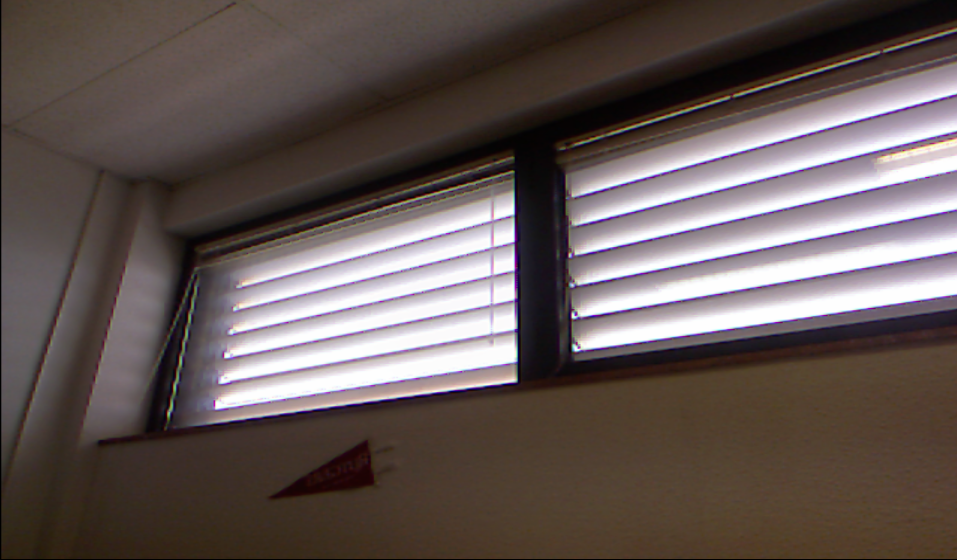}}
		 {\includegraphics[height=0.7in]{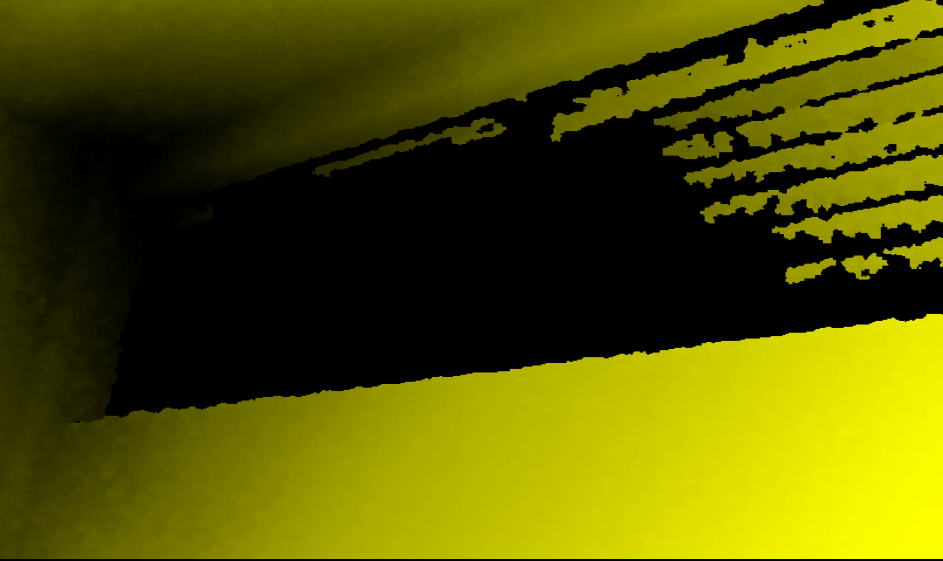}}
         {\includegraphics[height=0.7in]{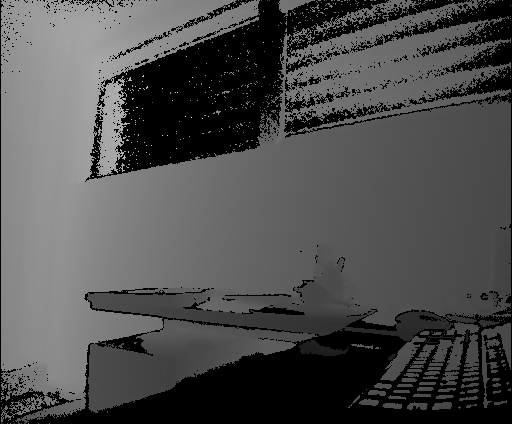}}
		
		\caption{Specular materials.}
    \end{subfigure}\\
     \begin{subfigure}[b]{0.5\textwidth}
         {\includegraphics[height=0.7in]{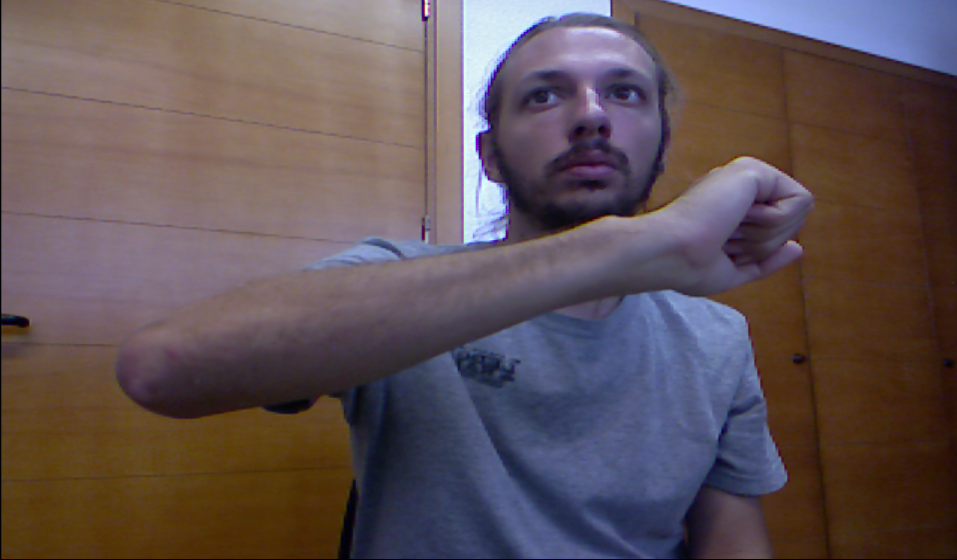}}
		 {\includegraphics[height=0.7in]{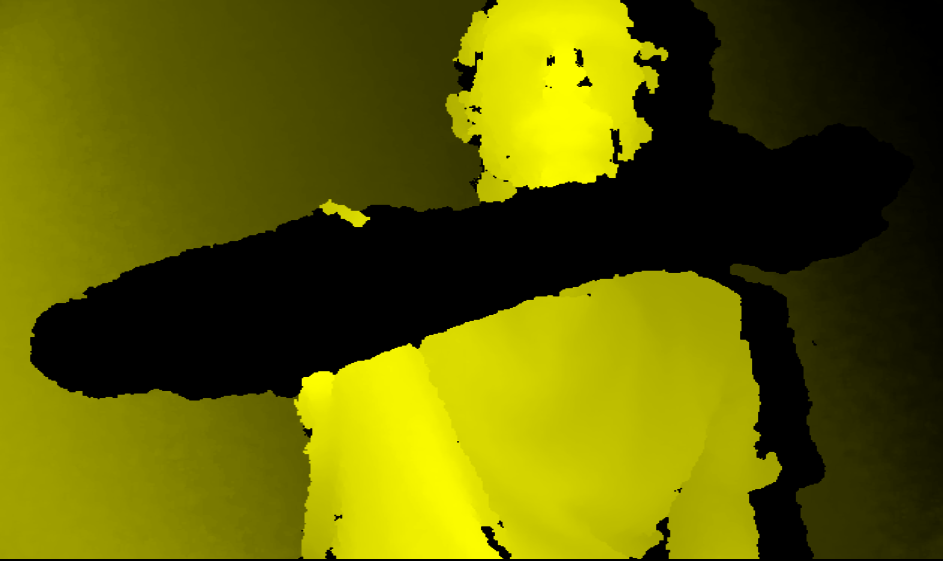} }
         {\includegraphics[height=0.7in]{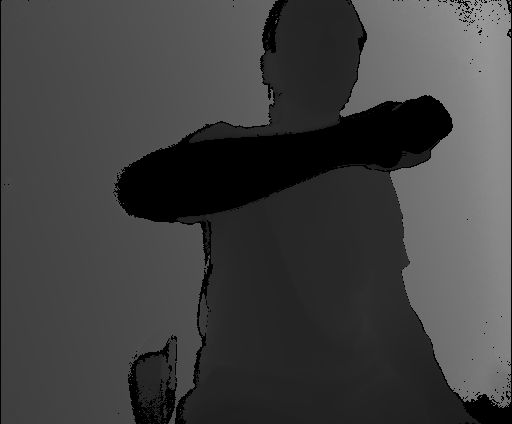}}
		\caption{Near objects.}
    \end{subfigure}\\
        \begin{subfigure}[b]{0.5\textwidth}
         {\includegraphics[height=0.7in]{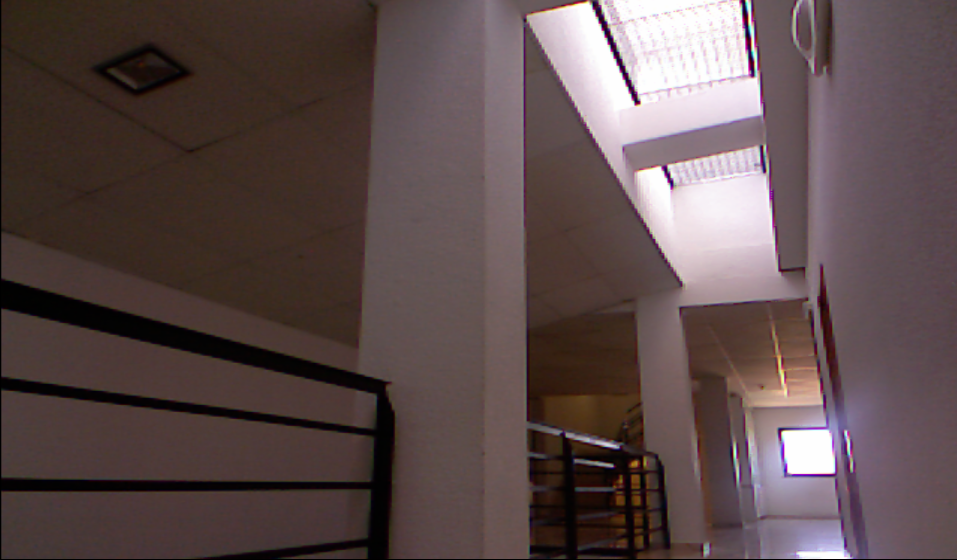}}
		 {\includegraphics[height=0.7in]{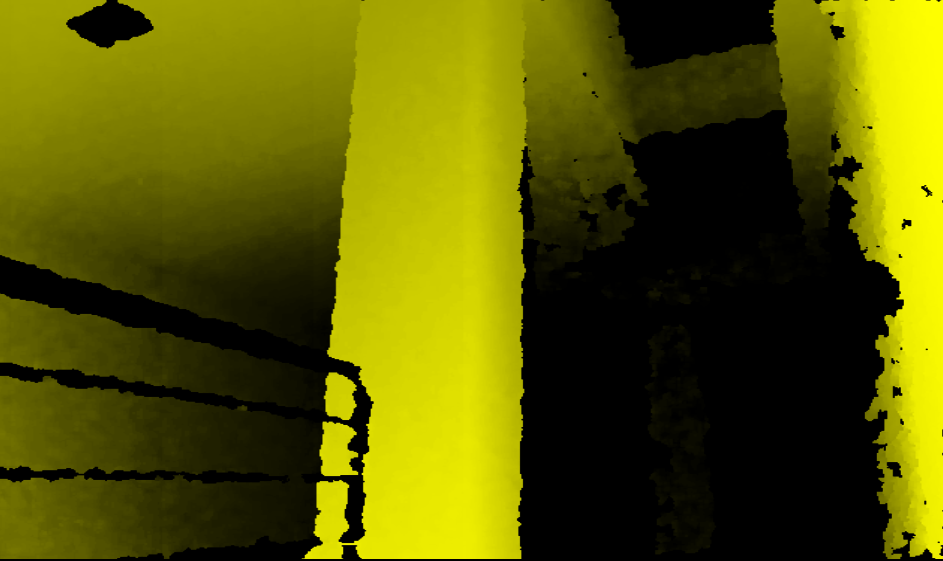} }
         {\includegraphics[height=0.7in]{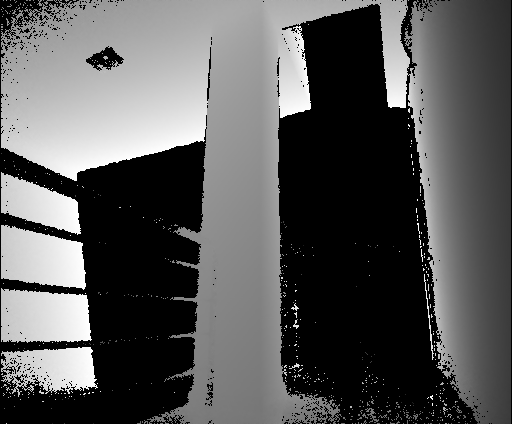}}
		\caption{Remote parts of the scene.}
    \end{subfigure}
     \begin{subfigure}[b]{0.5\textwidth}
         {\includegraphics[height=0.7in]{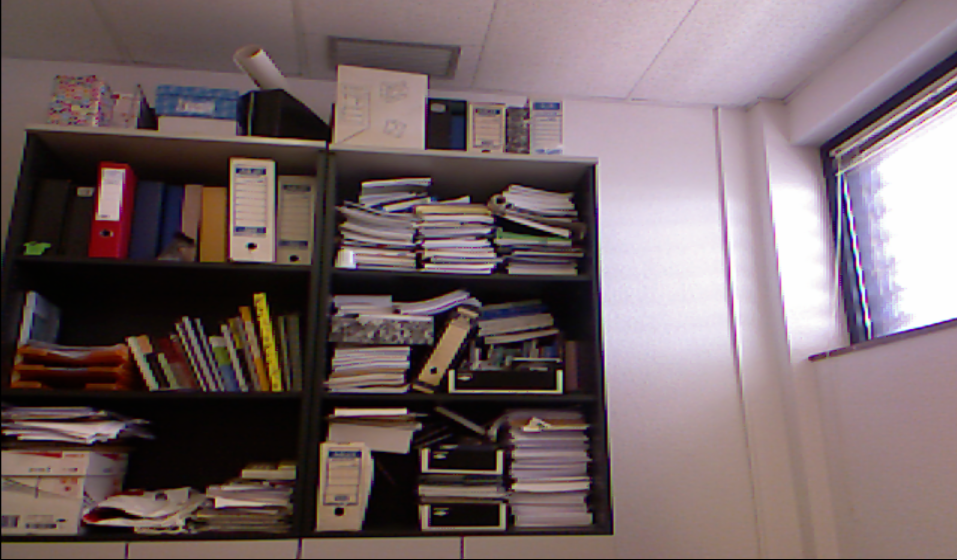}}
		 {\includegraphics[height=0.7in]{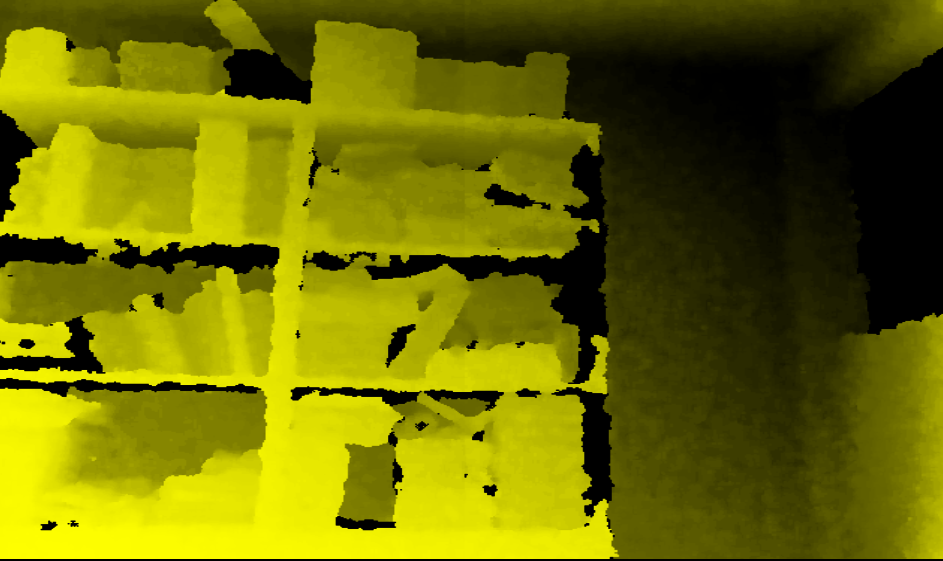} }
         {\includegraphics[height=0.7in]{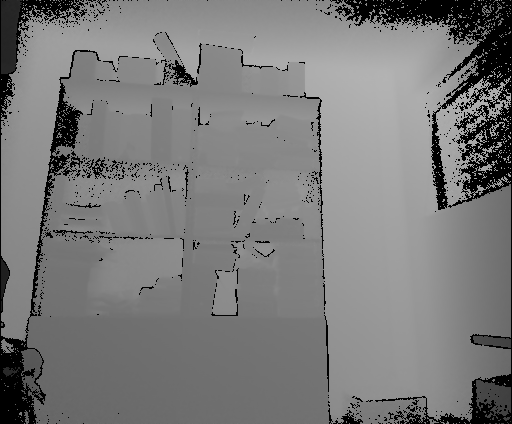}}
		\caption{Non reachable areas.}
    \end{subfigure}\\
    \begin{subfigure}[b]{0.5\textwidth}
         {\includegraphics[height=0.7in]{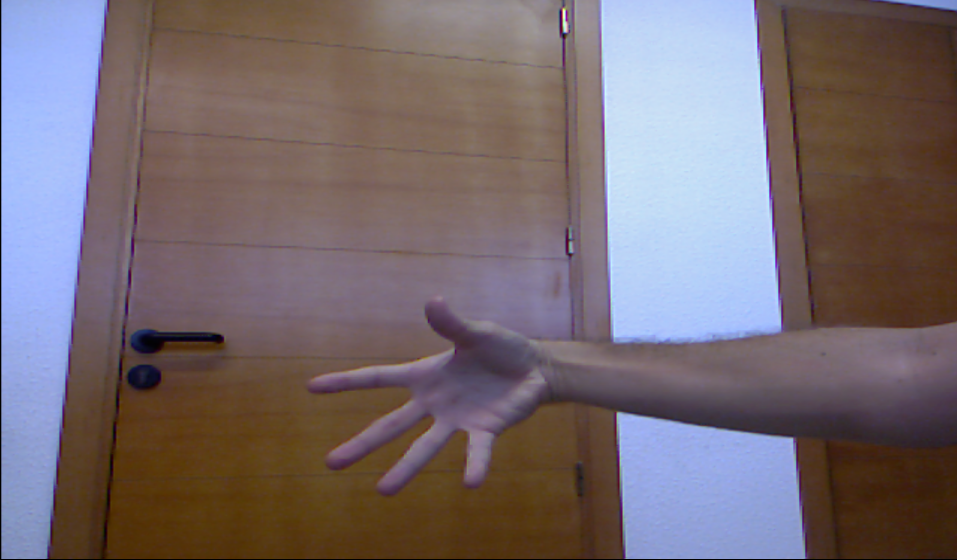}}
		 {\includegraphics[height=0.7in]{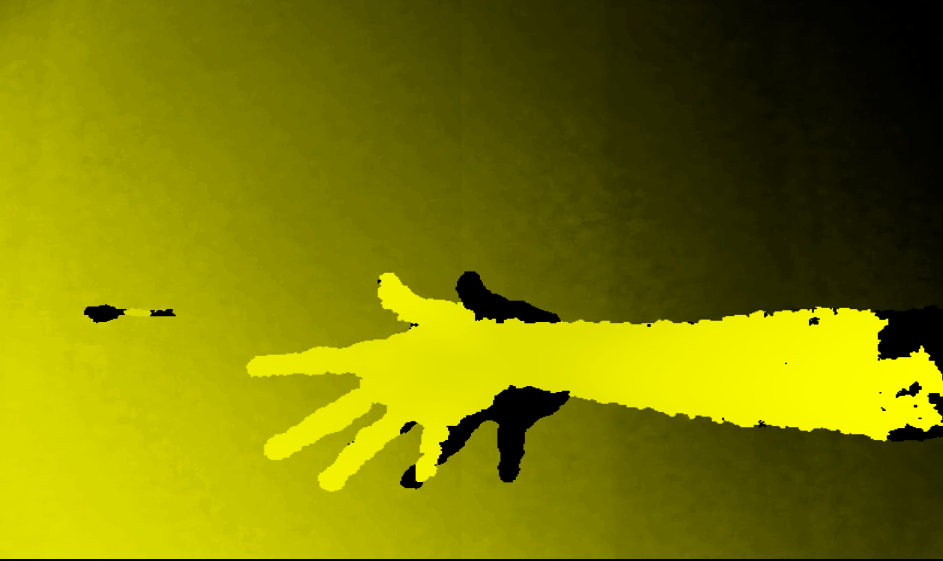} }
         {\includegraphics[height=0.7in]{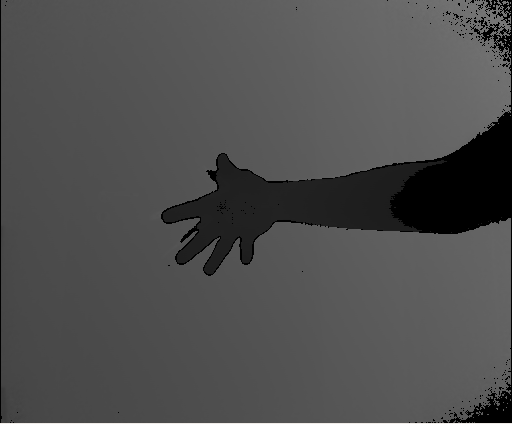}}
		\caption{Shadows.}
    \end{subfigure}\\
    \caption{Challenges of depth data: Each row illustrate a different problem. Second column corresponds to depth channel of Structured light sensor and third column corresponds to to depth channel of Time-of-flight sensor observations. Black regions in depth images corresponds Absent Depth Observations.}
    	\label{fig_depth_problems}
\end{figure}

Independently of which technology is used, depth data estimated by these devices suffer from several problems, which we describe here. 
Fig.~\ref{fig_depth_problems} illustrates examples of these problems. 

\begin{enumerate}
\item \textbf{Depth camouflage} (Fig.~\ref{fig_depth_problems}-a): Due to sensor sensitivity, when the foreground and background are close in depth, the sensor gives the same depth data values. This makes it hard to segment the foreground from the background based on depth. 

\item \textbf{Specular materials} (Fig.~\ref{fig_depth_problems}-b): Rays from a single incoming direction are reflected back in a single outgoing direction without causing the diffusion needed to obtain depth information.  

\item \textbf{Near objects} (Fig.~\ref{fig_depth_problems}-c): Sensors have minimum depth specifications. Due to the proximity of the foreground objects, the sensor is unable to measure depth. Typically, both structured light sensors and time-of-flight sensors have a depth limit of 0.5 meters. 

\item \textbf{Remote parts of the scene} (Fig.~\ref{fig_depth_problems}-d): Sensors have maximum distances at which they can detect depth. Parts of the scene farther from this distance appear as gaps in depth images.

\item \textbf{Non reachable areas} (Fig.~\ref{fig_depth_problems}-e): Depending on the imaging geometry and the sensor position, parts of the background may be occluded. This makes the sensor unable to estimate the depths at these locations.

\item \textbf{Shadows} (Fig.~\ref{fig_depth_problems}-f): Foreground objects block the active light emitted by the sensor from reaching the background, which causes shadows to be cast on the background. Thus the sensors are unable to estimate the depth at these blocked regions. Therefore, RGBD sensors exhibit two different types of shadows: visible-light shadows in the RGB channels, and IR shadows in the depth channel. These two different types of shadows are different in their geometries and in their spatial extents in the image.
\end{enumerate}

When depth cannot be measured at a given pixel, as in cases 2 to 6 above, the sensor returns a special non-value code to indicate its inability to measure depth. Such pixels appear as holes in the images with absent depth value. In this paper we denote these pixels as Absent Depth Observations (ADO).

In the next section, we define a scene model that manages these data issues using both color and depth information in a unified way.

\section{Non-parametric Scene Model}
\label{sec:BackgroundModel}

\subsection{Statistical background model}
Our model is based on recent scene information. Given the last $n$ observations of a pixel, denoted by 
$\mathbf{x}_i$ , $i = 1, \ldots, n$ in the d-dimensional observation space $ \mathbb{R}^d$, which enclose the sensor data values, it is possible to 
estimate the probability density function (pdf) of each pixel with respect to all previously observed values~\cite{Elgammal2000, 
scott1992multivariate}

\begin{equation} \label{eq:multydensity}
P(\mathbf{x}) = \frac{1}{n}|\textbf{H}|^{-\frac{1}{2}} \sum\limits_{i=1}^{n}K(\textbf{H}^{-\frac{1}{2}}( \mathbf{x}- \mathbf{x}_i)) \ ,
\end{equation} 

\noindent where $K$ is a multivariate kernel, satisfying $\int K(x)dx = 1$ and  $K(u)$ $\geq$ 0. \textbf{H} is the bandwidth matrix, which 
is a symmetric positive \textit{d$\times$d}-matrix.

The choice of the bandwidth matrix \textbf{H} is the single most important factor affecting the estimation accuracy because it controls the amount and orientation of smoothing induced~\cite{wand1994kernel}. Diagonal matrix bandwidth kernels allow different amounts of smoothing in each of the dimensions and are the most widespread due to computational reasons~\cite{dbandwith}. The most commonly used kernel density function is the Normal function, in our approach $N(0,\textbf{H})$ is selected

\[ \textbf{H} =\left( \begin{array}{cccc}
\sigma_1^2 & 0 & \cdots & 0 \\
0       & \sigma_2^2 & \cdots & 0 \\
\vdots  & \vdots     & \ddots & \vdots \\  
0 & 0 & \cdots& \sigma_d^2 \end{array} \right) \]

\noindent where $\sigma_i^2$ is bandwith of the kernel in the \textit{i-th} dimension, \textit{i.e.} independence between the different channels is assumed.
The final probability density function can be written as

\begin{equation} \label{eq:elgammalD}
P(\mathbf{x}) = \frac{1}{n} \sum\limits_{i=1}^{n} \prod\limits_{j=1}^{d}\frac{1}{\sqrt{2\pi\sigma_j^2}}e^{-\frac{1}{2}\frac{(x_{j} - x_{ij})^2}
{\sigma_j^2}} \ .
\end{equation}

\noindent Given this estimate at each pixel, a pixel is considered foreground if its probability is under a certain threshold.


To estimate the kernel bandwidth $\sigma_j^2$ for the $jth$ dimension for a given pixel, similar to~\cite{Elgammal2000}, we compute the median absolute deviation over the data for consecutive values of the pixel. That is, the median, \textit{$m_j$}, of 
each  consecutive pair 
in the data is calculated independently for each dimension. Because we are measuring deviations between two 
consecutive values, each  pair 
usually comes from the same local-in-time distribution and only few pairs are expected to 
come from cross distributions. Assuming that this local in-time distribution is Normal $N(\mu, \sigma)$ 
then the deviation 
is Normal $N(0, 2\sigma_j^2)$. Therefore, the standard deviation of the first distribution can be estimated as in

\begin{equation} \label{eq:bandwith}
\sigma_j = \frac{m_j}{0.68\sqrt{2}} \ .
\end{equation}

To create a fast implementation of the algorithm, the probability is estimated given the pixel value difference and the Kernel function bandwidth using a precalculated lookup table.

Prior to the use of this scene model, it is necessary to perform a training stage in which models of color and depth information are learned and the bandwidth of each channel used is calculated at each pixel.

\subsection{Depth data modeling}
\label{sec:depthdatarepresentation}
The scene model cannot be applied in a standard way because the sensor’s ADO requires a special treatment in which depth is treated as just a fourth channel, in addition to RGB. These ADOs can introduce errors into our model as well as into any typical background model. A pixel can be ADO throughout the sequence or switch in a random manner between ADO and a valid value.

We distinguish two categories of ADO:
\begin{itemize}
\item ADOs provoked by the scene's physical configuration. They belong to the background, even in the absence of foreground objects. These include specular background materials, remote parts of the scenes and non-reachable areas.
\item ADOs caused by the foreground objects. These include nearby objects, specular foreground objects and shadows.
\end{itemize}

We want to differentiate these two classes of ADO pixels (see  Fig.~\ref{fig_depth_problems}). We propose a probabilistic model, which we call the ADO model. The probability of a pixel being ADO and belonging to the background model is denoted by $P_{A}$. This probability is calculated for the depth component of each pixel $D$. 
The ADO model is updated for each pixel during the training stage. $P_{A}$ is calculated recursively as follows:

\begin{eqnarray}
P_{A}(D_0) &~=~& 0\\ \nonumber 
P_{A}(D_t) &~=~& \alpha*mask_t +(1-\alpha)*P_{A}(D_{t-1}) \ ,
\end{eqnarray}

\noindent where $\alpha$ $ \in [0,1] $ is the update rate, and $mask_t$ is a binary value corresponding to an ADO-mask, where each pixel have 
value of 1 if $D_t$  is ADO and 0 if not.

\begin{figure}[!htbp]
    \begin{subfigure}[b]{1.0\textwidth}
    
    		{\centering \includegraphics[width=1.7in]{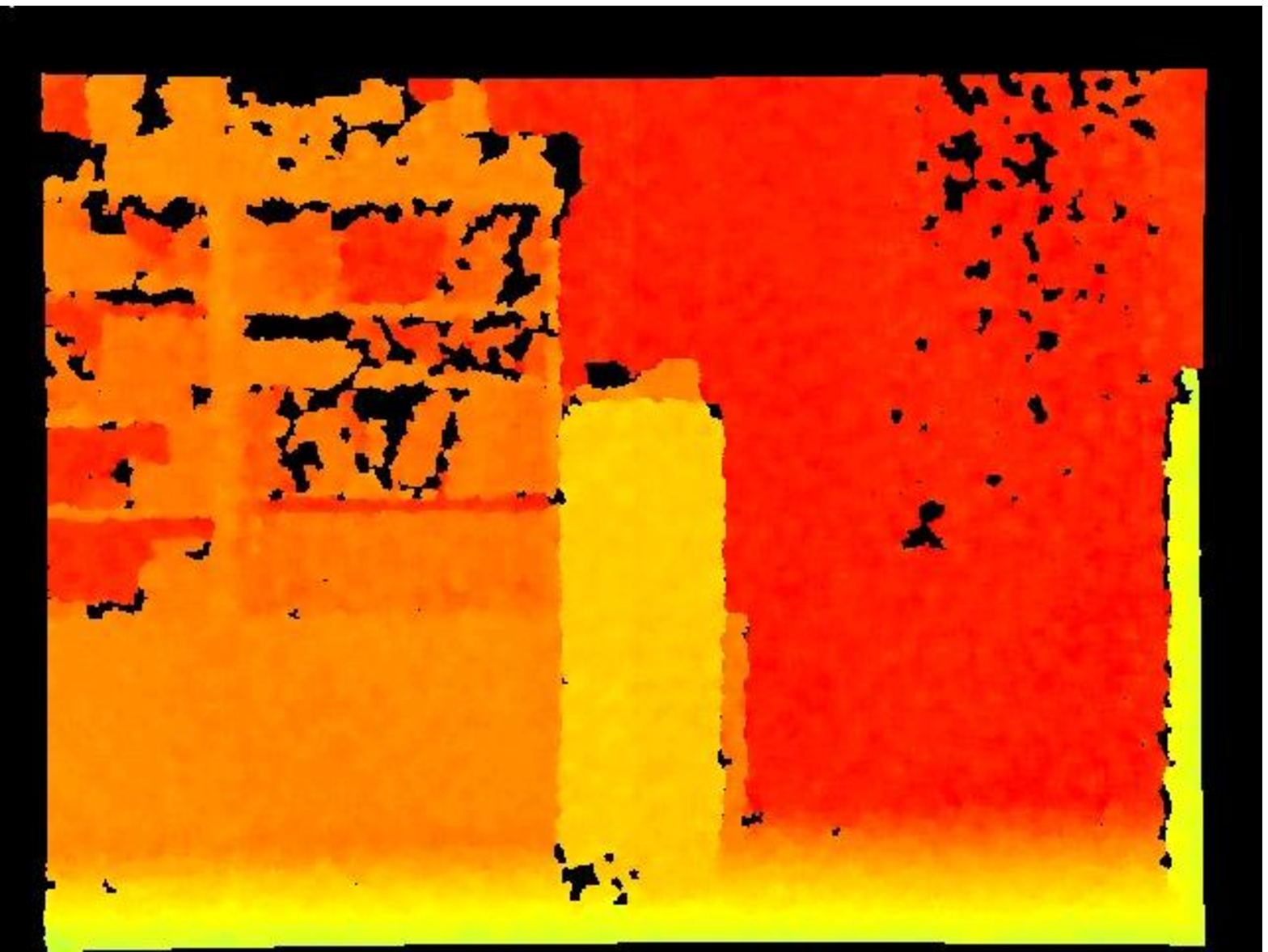}}
    		{\centering \includegraphics[width=1.7in]{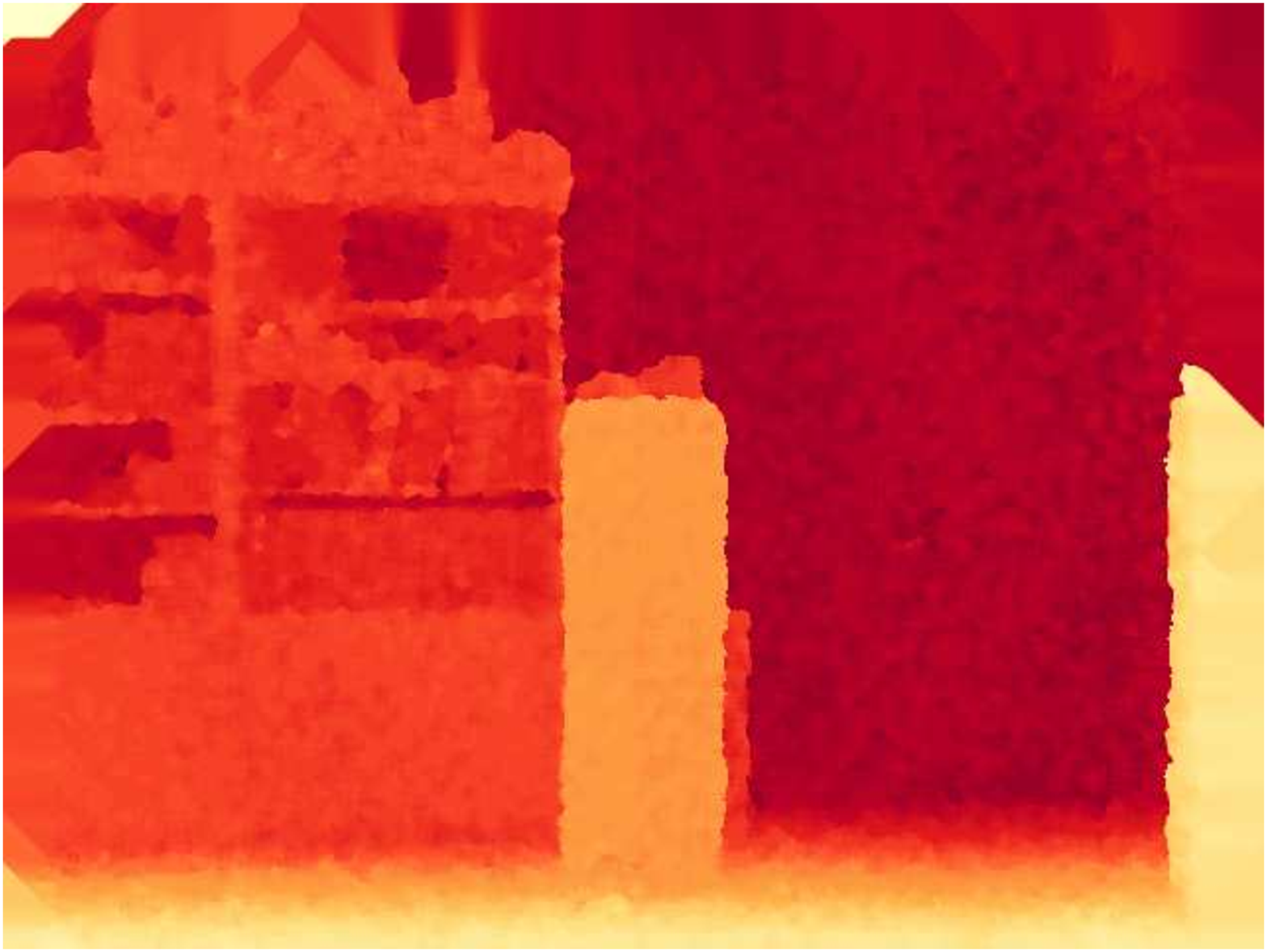}}
    \end{subfigure}\\
    \caption{The inpainting process is done over first training frame in order to overwrite undefined values. In next frames these values are 
    propagated under the undefined values model.}
    \label{fig_ADO}
\end{figure}

We try to avoid adding an ADO pixel to the background model. We selected a strategy based on overwriting the pixel with the previous 
one. Let $D_t$, $t = 1, \ldots, n$ be the $n$ recently sampled depth values at a given pixel, $D_t$ is calculated as follows:

\begin{equation}\label{eq:overwritten}
\setlength{\nulldelimiterspace}{0pt}
D_t=\left\{\begin{IEEEeqnarraybox}[\relax][c]{l's}
D_{t-1},& if $ D_t = ADO$\\
 D_t,& $otherwise$%
\end{IEEEeqnarraybox}\right. \ ,
\end{equation}

\noindent where for $D_0$, we use the inpainting strategy suggested by~\cite{Greff2012}, in which the initial image reconstruction algorithm of~\cite{telea2004image} tries to estimate the correct values for ADO regions (see Fig.~\ref{fig_ADO}).

\begin{figure}[!htbp]
    \begin{subfigure}[b]{1.0\textwidth}
    		{\includegraphics[width=3.5in]{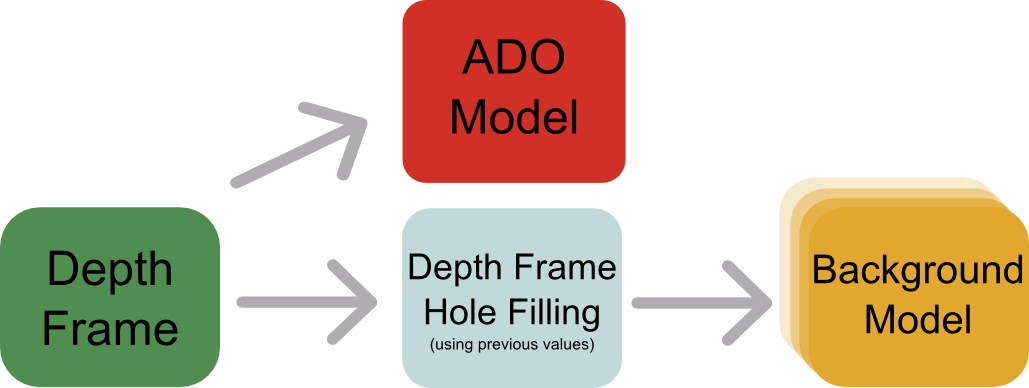}}
    \end{subfigure}\\
    \caption{Training step: In the first frame we apply an inpainting process. Following depth frames each ADO pixel is overwritten by 
    a previous value. Once this process is done pixels are added to background model.}
    \label{fig_training}
\end{figure}

The ADO model is calculated for each pixel during the training phase. Fig.~\ref{fig_training} depicts the ADO model's computation process. 
Pixels with $P_{A}$ higher than a threshold $\theta$ are overwritten with a previous value and incorporated into the background 
model. The other pixels remain undefined and are then classified as the undefined class.

\subsection{Background moving object detection}
\label{sec:sleepingObjects}
In real scenes, a background object can be moved. Such area should not be considered part of the background forever; therefore, the scene model has to adapt and understand that the scene layout may be physically updated~\cite{Cristani2010}.Typically, in background subtraction algorithms, the new background is incorporated into the model at a speed that depends on the corresponding update rate. By introducing depth data values, we present a new approach which permits instantaneous pixel classification.

The idea is based on the fact that if a new depth observation is located farther than the modelled values, this is probably because it became part of the background when some object was removed from the scene. To detect these changes, we compare the difference between this new observation and the previous observation with the background model, and we check whether that difference is larger than the modelled one for each pixel. The cumulative density function (cdf) over the absolute difference of two consecutive observations of a pixel allows us to formalize this idea.

Given the observations $D_1, \ldots, D_n$ the \textit{ith} component of vector \textbf{V} is defined as $V_i$ = $\mid D_i - D_{i-1}\mid$ , 
$\forall$ $ i = [2 \ldots n]$ for all k possible sensor values, k $\in$ \{0\ldots \textit{L}\}, where \textit{L} is the maximum number of depth levels.  

Then, we define $ P(k) = \frac{1}{n}\#\{V_i:V_i=k\}$, \\ and 
\begin{equation}
F_x(k)= \sum^k_{j=1} P(j) \ .
\end{equation}

Finally, given a new observation $D_t$ and the observations $D_1, \ldots, D_n$, we define the \textit{ith} component of evaluation set $C_{D_t}$ as the threshold to zero of the difference. That is 

\begin{equation}\label{eq:overwritten2}
\setlength{\nulldelimiterspace}{0pt}
C_{D_{t,i}}=\left\{\begin{IEEEeqnarraybox}[\relax][c]{l's}
 D_t - D_{i},& if $  D_t - D_{i} > 0$\\
 0 ,& $otherwise$%
\end{IEEEeqnarraybox}\right. \ ,
\end{equation}

\noindent $\forall$ $ i = [1 \ldots n]$. The evaluation function for background moving objects detection: 

\begin{equation}\label{eq:mod}
U(D_t) =  \frac{\sum_{\forall k \in C_{D_t}} F_x(k)}{n} \ .
\end{equation}

If $U(D_t)$ is higher than a predefined threshold, $\xi$, the pixel is considered part of the background. This detection is very relevant, so 
physical changes in the scene are detected when occurs. 

\section{Model update}
\label{sec:ModelUpdate}
In previous sections, we detailed how to detect foreground regions of a scene given a recent history of samples for each pixel. This model needs to be updated to properly respond to changes in the scene. Because the kernel bandwidth estimation requires all of the samples to be close in time, the update is performed using a first-in first-out queue: the oldest sample is discarded and a new sample is added to the model. 

Different updating strategies are used to keep the model updated. On one hand, color information tends to have quick variations due to shadows and varying luminance; therefore, we consider it an unstable model. On the other hand, depth information tends to be more stable.
 
\subsection{Color update}

The intensity distribution of the color information can change dramatically over very short periods of time~\cite{Elgammal2000}. For each frame, the color model is updated so the model can adapt very quickly to changes in the background process. A new observation is added to the model only if it is classified as a background sample. If a pixel is updated with the foreground color value, the error will be propagated over time and misclassification problems will appear. To avoid the adaptation of the model to the foreground object characteristics, a higher threshold is proposed to relax the condition and avoid updating pixels that are very close to belonging to the foreground.

\subsection{Depth update}

Unlike color, depth information represents a stable long-term description of the scene. Therefore, it is not necessary to update the model for each frame as pixel values do not change as fast as color values. Pixels detected as a part of a background moving object are automatically classified as background and their models are updated. In fact, updates to the depth model are highly related to physical changes in the real-world scene; therefore, pixels detected as background moving objects (see Section~\ref{sec:sleepingObjects}) are selected to be updated. In addition, the ADO model is updated for these pixels during this update phase.

\begin{figure}[!t]
\linespread{0.1} 
\footnotesize

\begin{mdframed}
Let define an observation $\mathbf{x} = \{r, g, D\}$, therefore d = 3. $\theta$, $\gamma$ and $\xi$ are constant values over all algorithm, where: $\gamma = 10^{-8}$, $\theta = 0.0050$ and $\xi = 0.6$.

\begin{center}
\textbf{Training stage} 
\end{center}


Initialization step:
\begin{itemize}
\item Image inpainting algorithm to compute $D_0$.
\item $P_A(D_0) = 0$.
\end{itemize}
 
Let ($\mathbf{x}_1$, \ldots, $\mathbf{x}_i$ , $\mathbf{x}_n$ ) be the observations used for modelling the scene for each $D_i \in \mathbf{x}_i$ \bigskip

\hspace{3.5 mm}Depth treatment: \\

\begin{itemize}

\item Compute the ADO-model: $P_{A}(D_i)  =\alpha*mask+(1-\alpha)*P_{A}(D_{i-1})$.		
\item \textit{If} $D_i$ is ADO-pixel \textit{then} $\hat{D}_i$ = $\hat{D}_{i-1}$ \textit{else} $\hat{D}_i = D_i$.

\item Substitute $\hat{D}_i$ for ${D}_i$ in  $\mathbf{x}_i$. 

\end{itemize}

\hspace{3.5 mm}Calculate the kernel bandwidth for each dimension \textit{d}. \\

\begin{itemize}
\item  $\forall j \in [1..d]$ compute the median absolute deviation $m_j$ for consecutive values of observations:
	\begin{center}
		$m_j = \mid x_{ji} - x_{ji+1} \mid  \forall$ \textit{i} $\in [1 \ldots t]$.
	\end{center}

\item Calculate de bandwith $\sigma_j = \frac{m_j}{0.68\sqrt{2}}$.

\end{itemize}

\begin{center}
\textbf{Segmentation stage} 
\end{center}
Let $\mathbf{x}_t$ be a new observation
\bigskip

\hspace{3.5 mm}Depth treatment:

\begin{itemize}
\item \textit{If} $D_t$ is $ADO$-pixel \textit{then} $\hat{D}_t$ = $\hat{D}_{t-1}$ \textit{else} $\hat{D}_t = D_t$.

\item Measure the probability of a pixel being part of a background moving object:
\begin{center}

$U(\hat{D}_t) =  \frac{1}{n}\sum_{\forall k \in C_{\hat{D}_t}}  F_x(k)$.
\end{center}
\item \textit{If} $U(\hat{D}_t) > \xi$ \textit{then} $\mathbf{x}_t$ $\in$ \textbf{background}.
\end{itemize}

\hspace{3.5 mm}Calculate the probability of a pixel being background for all dimensions, $d$:

\begin{itemize}

\item $
P(\mathbf{x}_t) = \frac{1}{n} \sum\limits_{i=1}^{n} \prod\limits_{j=1}^{d}\frac{1}{\sqrt{2\pi\sigma_j^2}}e^{-\frac{1}{2}\frac{(x_{nj} - x_{ij})^2}{\sigma_j^2}}$.

\item \textit{If} $P(\mathbf{x_t}) < \gamma$ \textit{then} $\mathbf{x}_t$ $\in$ \textbf{foreground} \textit{else} $\mathbf{x}_t$ $\in$ \textbf{background}.
\item Classify $\mathbf{x}_t$ $\in$ as \textbf{undefined} \textit{If} $D_t $ is $ADO$-pixel \textit{and} $P_{A}({D}_t) < \theta$.
\end{itemize}

\hspace{3.5 mm}Update background model:

\begin{itemize}
\item \textit{If} $\mathbf{x}_t$ $\in$ \textbf{background} \textit{then} update color model.
\item \textit{If} $U(\hat{D}_t) > \xi$ \textit{then} update depth model.
\end{itemize}

\end{mdframed}

\caption{The generic scene modelling algorithm for RGBD devices.}
\label{fig_algorithm}
\end{figure}

\section{Generic Scene Modeling (GSM)}
\label{sec:Exp}
In this Section, we describe an experimental configuration of the scene modeling algorithm for evaluation purposes. Specifically, we explain the sensor color and depth inputs, including the algorithm parameters used. Finally, we define the generic scene model. In Fig.~\ref{fig_algorithm} the complete algorithm details are given.
\subsection{Depth input}
\label{sec:depthusage}
To evaluate the previously defined scene model algorithm, a Microsoft Kinect 1 sensor is used as an RGBD device. The device’s technology is based on structured light. The image processor of the sensor uses the relative positions of the dots in the pattern to calculate the depth displacement at each pixel position in the image~\cite{Khoshelham2012}. We use the sensor’s continuous raw depth information, where D in [650, 1500] which corresponds to a valid depth range between 0.5 and 4.5 meters. In addition, for this sensor, all ADOs have the same value of 650.

\subsection{Color input}
\label{sec:colorusage}
Usually color information is useful for suppressing shadows from detection by separating color information from lightness information. To construct a robust algorithm that is independent of illumination variations, we separated color information from luminance information using a non-luminance dependent color space. Then, color is defined as a combination of luminance, hue and saturation.  Chromaticity is the description of a color ignoring its luminance, and it can be described as a combination of hue and saturation. Given the device's three color channels \textit{R, G, B}, the chromaticity coordinates \textit{r, g} and \textit{b} are: $r = {R}/{(R+G+B)}$, $g = {G}/{(R+G+B)}$, $b = {B}/{(R+G+B)}$ where: $r+g+b = 1$~\cite{levine1985vision}. In our model we use two dimensions: $r$ and $g$.

\section{Evaluation}
\label{sec:evaluation}

The evaluation is performed using two implementations of the proposed GSM algorithm. GSM$_{{UB}}$ is used if undefined pixels are considered as background and GSM$_{{UF}}$ is used if undefined pixels are considered as foreground. We used two datasets to perform the tests.First, we perform the comparison using a dataset that emphasizes camouflage and shadows problems. We selected this dataset because it facilitates comparison with other background subtraction algorithms that use both color and depth information~\cite{Camplani2014}.

Second, a new RGBD sequence dataset is built inspired by the WallFlower~\cite{Toyama1999} dataset, one of the most widely used 
color-based background subtraction datasets. This dataset is built to test all background 
subtraction issues described in Wallflower besides the new depth challenges described in Section~\ref{sec:challengesOfDepthData}.

Different metrics are used to measure the algorithm's performance, in each test. All are based on \textbf{True Positives 
(TP)}, which count the number of correctly detected foreground pixels; \textbf{False positives (FP)}, which count the number of background pixels incorrectly classified as foreground; \textbf{True negatives (TN)}, which count the number of correctly classified background pixels;  and \textbf{False negatives (FN)}, which count the number of foreground pixels incorrectly classified as background.

\subsection{Camplani Dataset}

\begin{figure}[t]
   \begin{subfigure}[b]{0.50\textwidth}
    \begin{subfigure}[b]{0.50\textwidth}
    		{\includegraphics[width=0.450\linewidth]{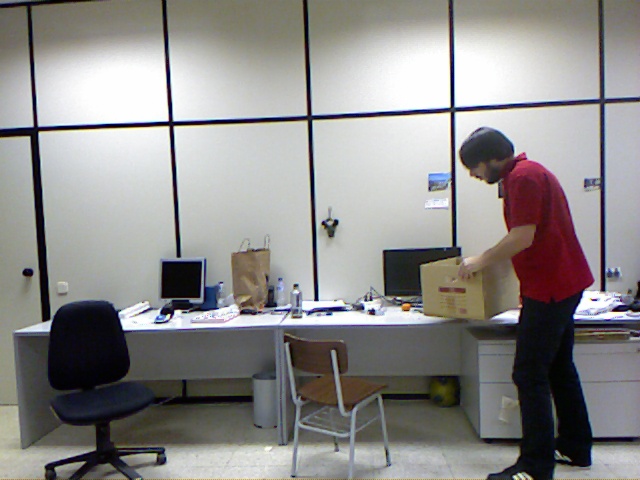}}
    		{\includegraphics[width=0.450\linewidth]{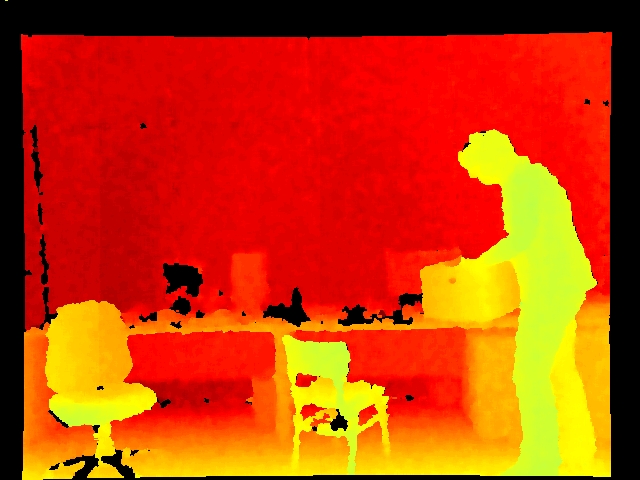}}
         \caption{GenSeq sequence}
    \end{subfigure}
    \hfill
     \begin{subfigure}[b]{0.50\textwidth}
         {\includegraphics[width=0.45\linewidth]{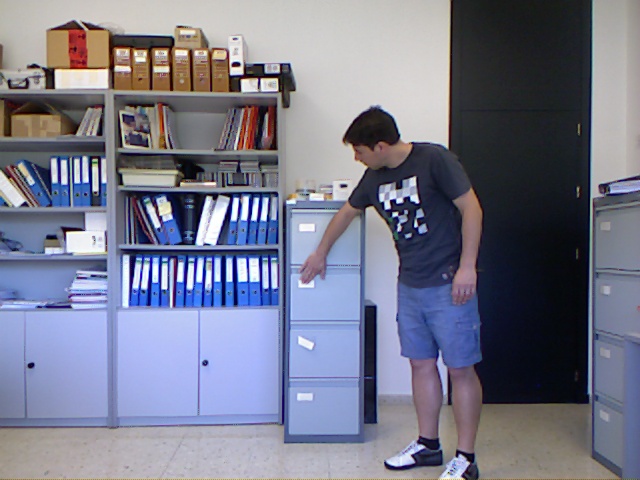}}
         {\includegraphics[width=0.45\linewidth]{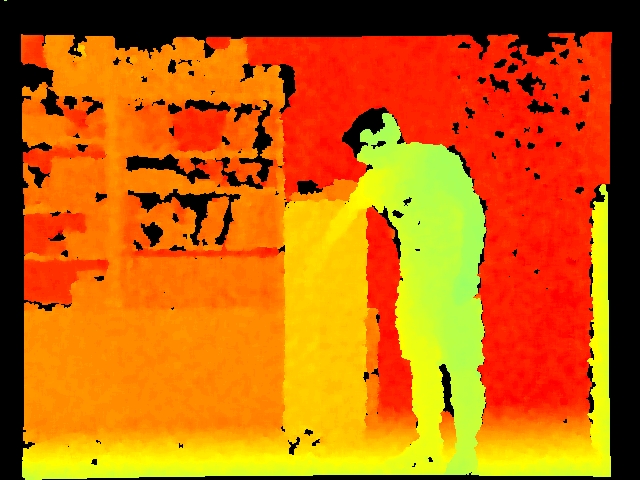}}
		\caption{DCamSeq sequence}
    \end{subfigure}
    \end{subfigure}
    \begin{subfigure}[b]{0.5\textwidth}
     \begin{subfigure}[b]{0.5\textwidth}
         {\includegraphics[width=0.45\linewidth]{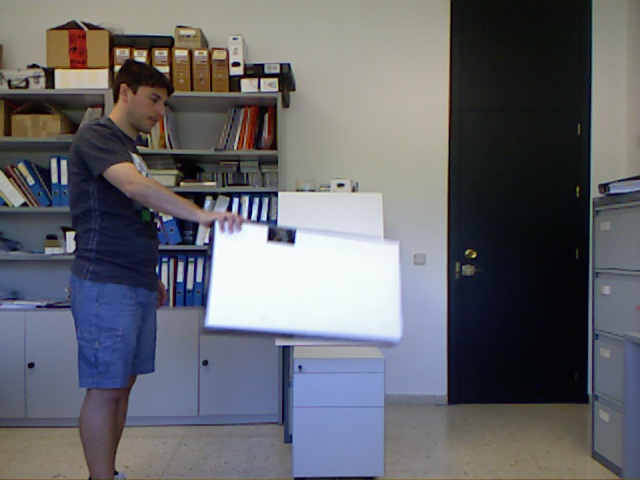}}
         {\includegraphics[width=0.45\linewidth]{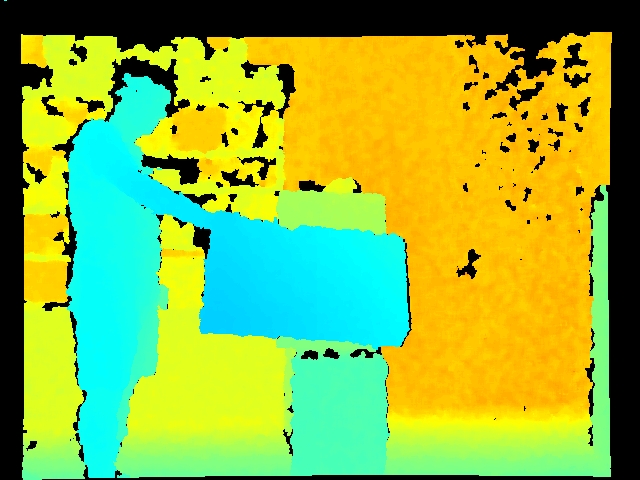}}
		
		\caption{ColCamSeq sequence}
    \end{subfigure}
  	\hfill
     \begin{subfigure}[b]{0.5\textwidth}
        {\includegraphics[width=0.45\linewidth]{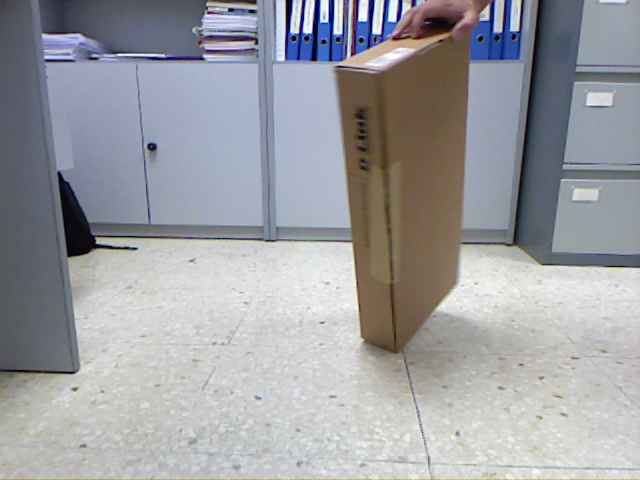}}
        {\includegraphics[width=0.45\linewidth]{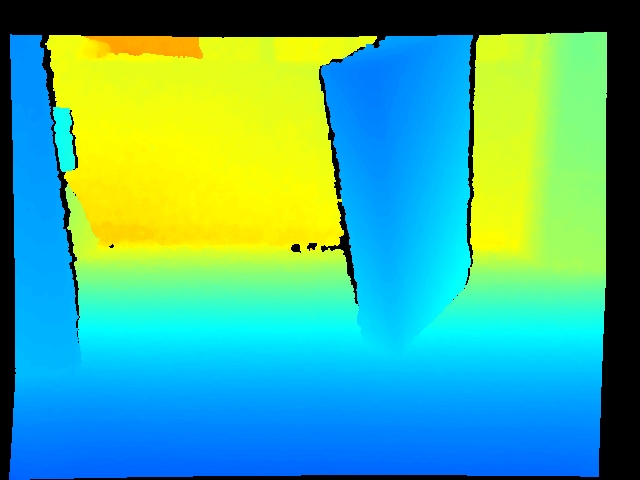}}
		\caption{ShSeq sequence}
    \end{subfigure}
    \end{subfigure}
    \caption{Color and depth frames examples for each sequence of Test I.}
    \label{fig_TestI}
\end{figure}

In~\cite{Camplani2014}, the authors presented a six-video RGBD dataset with hand-labelled ground truth (see Fig.~\ref{fig_TestI} and 
Table~\ref{testI}). The authors compared eight different background subtraction algorithms: the Camplani algorithm: CL$_{{W}}$; two weak classifiers,CL$_{{C}}$ and CL$_{{D}}$, defined in their paper; four different implementations of mixture of Gaussians and ViBe. To perform the evaluation, they used seven measures: \textbf{FN}; \textbf{FP}; \textbf{Total error (TE)}, the total number of misclassified pixels normalized with respect to the image size; a \textbf{Similarity measure (S)}, which is a non-linear measure that fuses FP and FN; and a \textbf{Similarity measure in object boundaries (S$_{B}$)}. S is close to 1 if detected foreground regions correspond to the real ones; otherwise its value is close to 0. $S_B$ is calculated similarly to S, but considering only the regions of the image surrounding the ground-truth object boundaries of 10 pixel width. Finally, two different metrics are used to rank the accuracy of the analyzed algorithms. \textbf{RM} ranks each method for each performance metric for one sequence. \textbf{RC}, a global ranking of the algorithms across different sequences, is the mean of \textbf{RM} for each method across all of the sequences.

\begin{figure}[htpb]
\includegraphics[width=3.5in]{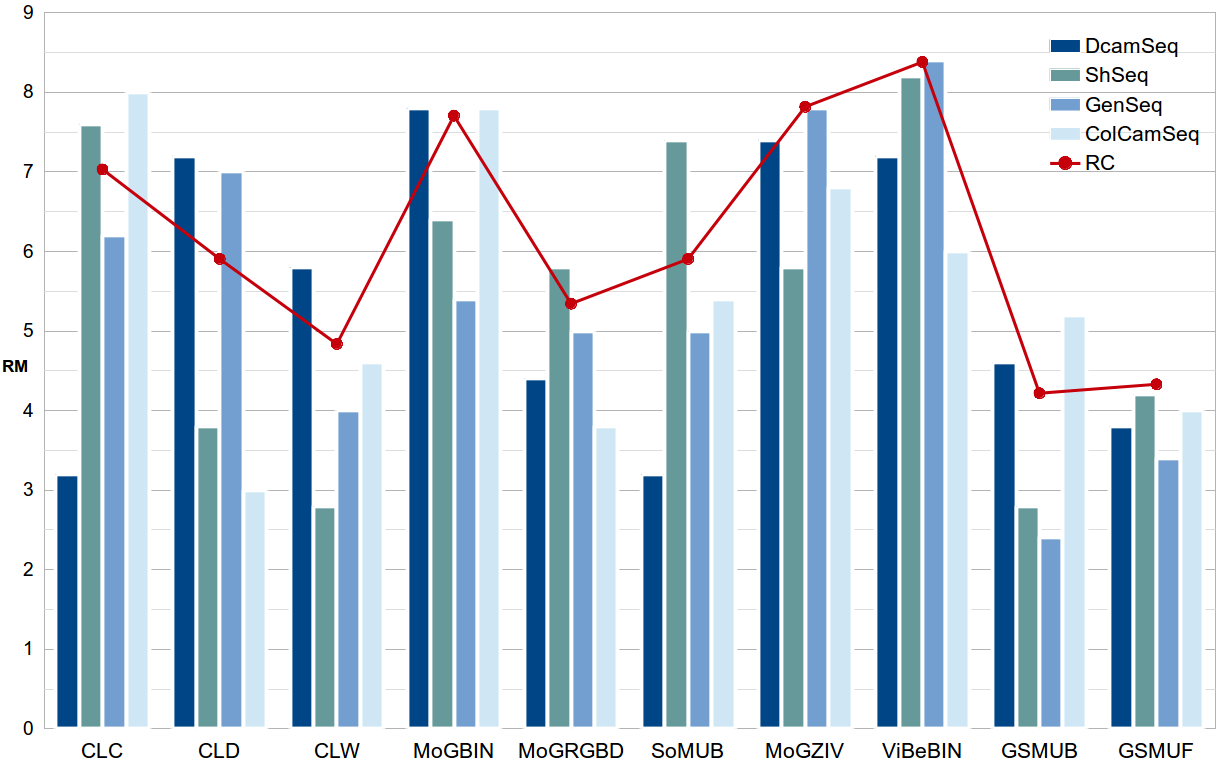}
\caption{Camplani dataset simulation Results (\textbf{RM}). It can be found results of four different sequences for all tested algorithms and final comparisons. As 
GSM$_{{UF}}$ and  GSM$_{{UB}}$ are not the best for each sequence, are the most regular ones as it can be seen with the \textbf{RC} line, the lower the better.}
\label{fig_sim_camplani}
\end{figure}

In Fig.~\ref{fig_sim_camplani}, global results are depicted. For each algorithm the ranking of each sequence is shown (RM). Finally, the RC classification is performed to establish a global result. Both GSM and CL$_{{W}}$ have the best results according to the RC ranking. To understand the global results, we analyze the performances of both algorithms for each sequence.

\begin{table} [!t]
\centering
\footnotesize
\renewcommand{\arraystretch}{1.3}
\centering
\caption{Characteristics of evaluated sequences from~\cite{Camplani2014} dataset.}

\begin{tabular}{C{1.3cm}C{1.3cm}C{1.4cm}C{1.5cm}C{1.3cm}}
\hline 
\textbf{Sequence} & \textbf{GenSeq} & \textbf{DCamSeq} & \textbf{ColCamSeq} & \textbf{ShSeq} \\ 
\hline 
\textbf{Number of frames} & 300 & 670 & 360 & 250 \\ 
\hline 
\textbf{Number of ground truth frames} & 39 & 102 & 45 & 25 \\ 
\hline 
\textbf{Test Objective} & General scenes & Depth camouflage & Color camouflage & Shadows \\ 
\hline 
\end{tabular} 
\label{testI}
\end{table}

The results in Table~\ref{Camplani_GenSeq} show that GSM$_{{UF}}$  has higher FN due to the classification of the entire undefined 
pixels-class as foreground. GSM$_{{UB}}$ achieves best results. In addition, both of our proposed solutions achieve better values in contours (S$_{{B}}$) than CL$_{{W}}$. 

The sequence \textit{ColCamSeq} (see Table~\ref{Camplani_ColCamSeq}) gives results opposite to those of \textit{GenSeq}. GSM$_{{UB}}$ has higher FN due to the misclassification of all undefined pixels as background. Again, our method gives better results for both sequences in contours (S$_{B}$) and in the similarity measure ($S$) than the CL$_{{W}}$ algorithm.

Our algorithm also achieves the best results in depth camouflage situations (see Table~\ref{Camplani_DCamSeq}), due to the combination of two type of information, color and depth, in the same model.

The results in Table~\ref{Camplani_ShSeq} show that in shadows our method obtains higher values of FP  compared with the CL$_{{W}}$ algorithm. In other measurements the proposed method gives better results.

\begin{table*} [!t]
\centering
\footnotesize
\caption{Results for \textbf{GenSeq}. \textbf{FP}: False positives. \textbf{FN}: False negatives. \textbf{TE}: Total error. \textbf{S}: Similarity measure. \textbf{S$_{{B}}$}: Similarity measure in object boundaries.}
\begin{tabular}{cccccccccccc}
\hline
 &\multicolumn{2}{c}{\bfseries TE} & \multicolumn{2}{c}{\bfseries FN} & \multicolumn{2}{c}{\bfseries FP} & \multicolumn{2}{c}{\bfseries S} & \multicolumn{2}{c}{\bfseries S$_{{B}}$} & RM\\
\hline
 &\bfseries Avg & \bfseries Std & \bfseries Avg & \bfseries Std &\bfseries Avg & \bfseries Std &\bfseries Avg & \bfseries Std &\bfseries Avg & \bfseries Std\\
\hline\hline
CL$_{{W}}$ & 1.30 & 0.42& 1.49 & 0.002& 1.27 & 0.01 & 0.83&0.21&0.53 & 0.14 & 3.2 \\
\hline 
GSM$_{{UB}}$ & 1.38 & 0.56& 1.04 & 0.78 & 1.44	 & 0.66 & 0.83 &	0.2 & 0.78 & 0.11 & 2.6
\\
\hline 
GSM$_{{UF}}$ &1.3   & 0.52 &4.08 & 15.38 & 1.3 & 0.6	 & 0.83 & 0.2 & 0.78 & 0.14 &	 3.2
 \\
\hline
\end{tabular}

\label{Camplani_GenSeq}
\end{table*}

\begin{table*}[!t]
\centering
\footnotesize
\caption{Results for \textbf{DCamSeq}. \textbf{FP}: False positives. \textbf{FN}: False negatives. \textbf{TE}: Total error.
 \textbf{S}: Similarity measure. \textbf{S$_{{B}}$}: Similarity measure in object boundaries.}
\begin{tabular}{cccccccccccc}
\hline
 &\multicolumn{2}{c}{\bfseries TE} & \multicolumn{2}{c}{\bfseries FN} & \multicolumn{2}{c}{\bfseries FP} & \multicolumn{2}{c}{\bfseries S} & \multicolumn{2}{c}{\bfseries S$_{{B}}$} & RM\\
\hline
 &\bfseries Avg & \bfseries Std & \bfseries Avg & \bfseries Std &\bfseries Avg & \bfseries Std &\bfseries Avg & \bfseries Std &\bfseries Avg & \bfseries Std\\
\hline\hline
CL$_{{W}}$& 2.46& 1.82&32.21& 0.26& 0.66&0.01& 0.55&0.14&0.5&10.12& 6.2 \\
\hline 
GSM$_{{UB}}$ &1.74 & 1.7 &	20.45 &	10.73 & 0.46 & 1.57 & 0.64 & 0.17 & 0.54 & 0.14 &	3.8
 \\
\hline 
GSM$_{{UF}}$ &1.65 & 1.49 &	22.06 & 11.6	 & 0.61 &1.73 &	0.65 & 0.18 & 0.55 & 0.14 & 3.6 \\
\hline
\end{tabular}

\label{Camplani_DCamSeq}
\end{table*}

\begin{table*}[!t]
\centering
\footnotesize
\caption{Results for \textbf{ColCamSeq}. \textbf{FP}: False positives. \textbf{FN}: False negatives. \textbf{TE}: Total error. \textbf{S}: 
Similarity measure. \textbf{S$_{{B}}$}: Similarity measure in object boundaries.}
\begin{tabular}{cccccccccccc}
\hline
 &\multicolumn{2}{c}{\bfseries TE} & \multicolumn{2}{c}{\bfseries FN} & \multicolumn{2}{c}{\bfseries FP} & \multicolumn{2}{c}{\bfseries S} & \multicolumn{2}{c}{\bfseries S$_{{B}}$} & RM\\
\hline
 &\bfseries Avg & \bfseries Std & \bfseries Avg & \bfseries Std &\bfseries Avg & \bfseries Std &\bfseries Avg & \bfseries Std &\bfseries Avg & \bfseries Std\\
\hline\hline
CL$_{{W}}$ & 3.20& 2.77& 3.52&0.09&2.92&0.10&0.89&0.15&0.77&0.16& 4.8 \\
\hline 
GSM$_{{UB}}$ & 2.3 &2.26&7.1&14.5	&3.21& 6.3 &0.9&0.15	&0.52&	0.11 &	5.2
 \\
\hline 
GSM$_{{UF}}$ & 2.2 & 2.27 & 2.94 & 5.53 & 4.36  & 6.42 & 0.92 & 0.08 & 0.53 &0.09 & 4
 \\
\hline
\end{tabular}

\label{Camplani_ColCamSeq}
\end{table*}

\begin{table*}[!t]
\footnotesize
\caption{Results for \textbf{ShSeq}. \textbf{FP}: False positives. \textbf{FN}: False negatives. \textbf{TE}: Total error. \textbf{S}: Similarity measure. \textbf{S$_{{B}}$}: Similarity measure in object boundaries.}
\centering
\begin{tabular}{cccccccccccc}
\hline
 &\multicolumn{2}{c}{\bfseries TE} & \multicolumn{2}{c}{\bfseries FN} & \multicolumn{2}{c}{\bfseries FP} & \multicolumn{2}{c}{\bfseries S} & \multicolumn{2}{c}{\bfseries S$_{{B}}$} & RM\\
\hline
 &\bfseries Avg & \bfseries Std & \bfseries Avg & \bfseries Std &\bfseries Avg & \bfseries Std &\bfseries Avg & \bfseries Std &\bfseries Avg & \bfseries Std\\
\hline\hline
CL$_{{W}}$ & 0.81& 0.35& 1.60& 0.05&0.68& 0.02& 0.94&0.04&0.71& 0.07 & 2.80 \\
\hline 
GSM$_{{UB}}$ & 0.87 & 0.33 & 0.98 & 0.88 & 0.88 & 0.42 & 0.93 & 0.03 & 0.76 &0.06 &	3
 \\
\hline 
GSM$_{{UF}}$ &1.66 & 0.38 & 0.14 & 0.19 & 1.92 & 0.44 & 0.89 & 0.04 & 0.65 & 0.05 & 4.2
 \\
\hline
\end{tabular}

\label{Camplani_ShSeq}
\end{table*}

\subsection{GSM dataset}

\begin{figure}[!t]
    \centering
    \begin{subfigure}[b]{0.5\textwidth}
    		{\includegraphics[width=1.1in]{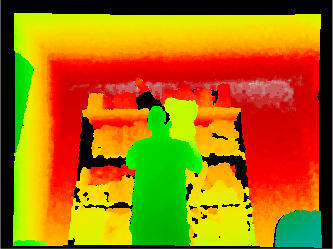}}
   		{\includegraphics[width=1.1in]{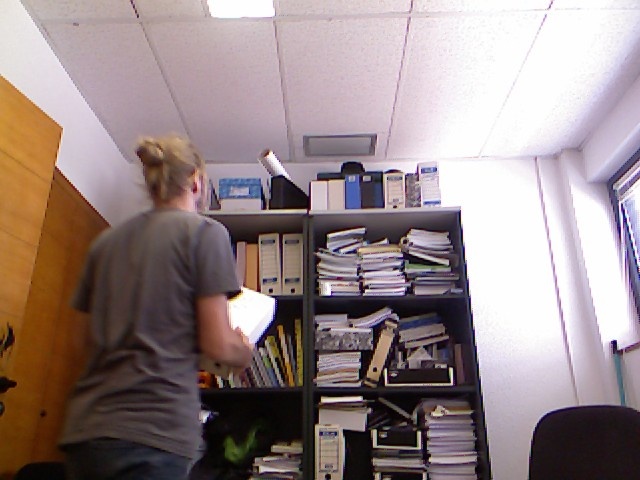}}
        {\includegraphics[width=1.1in]{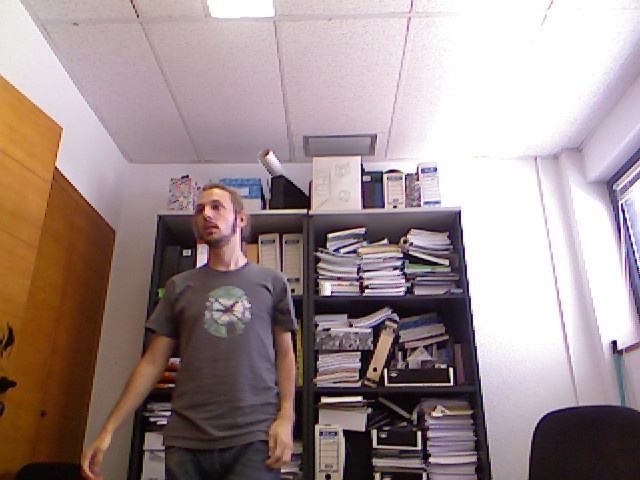}}
         \caption{\textbf{Color camouflage} sequence}
    \end{subfigure}\\
     \begin{subfigure}[b]{0.5\textwidth}
         {\includegraphics[width=1.1in]{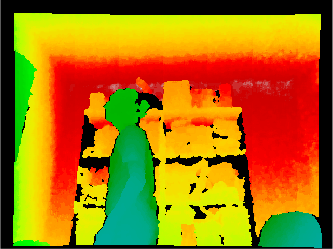}}
		 {\includegraphics[width=1.1in]{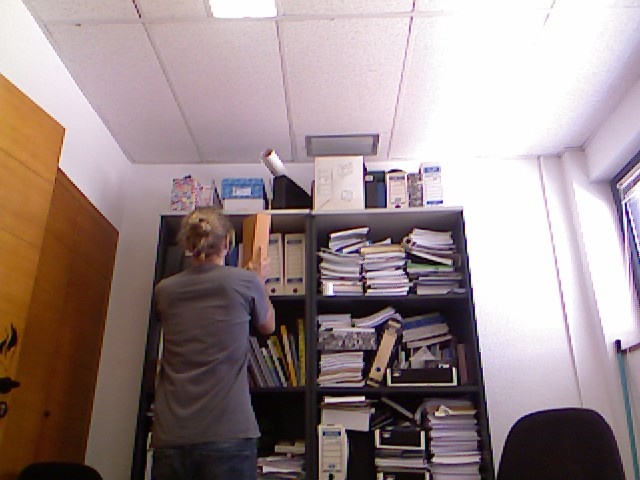} }
		{\includegraphics[width=1.1in]{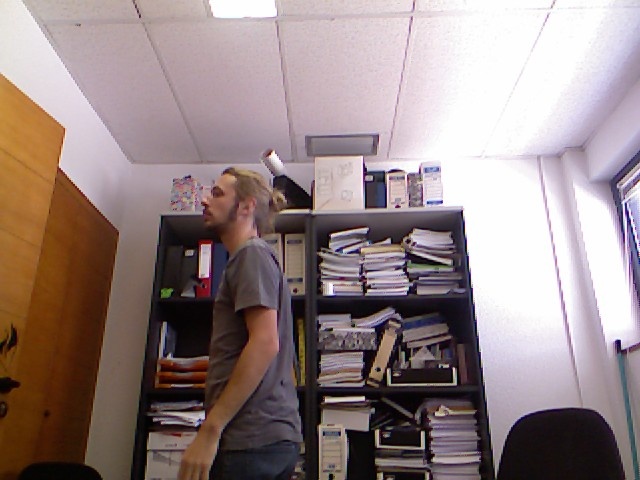} }
		\caption{\textbf{Depth camouflage} sequence}
    \end{subfigure}\\
     \begin{subfigure}[b]{0.5\textwidth}
         {\includegraphics[width=1.1in]{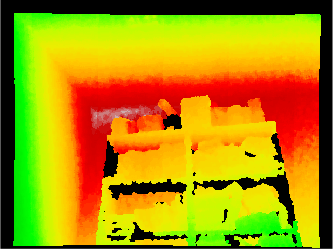}}
		 {\includegraphics[width=1.1in]{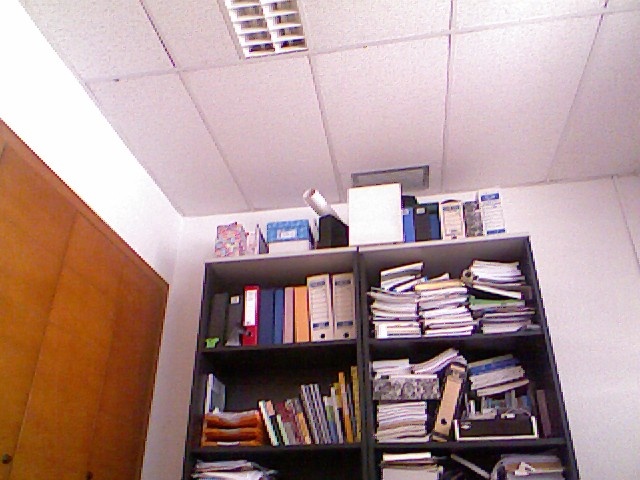} }
		{\includegraphics[width=1.1in]{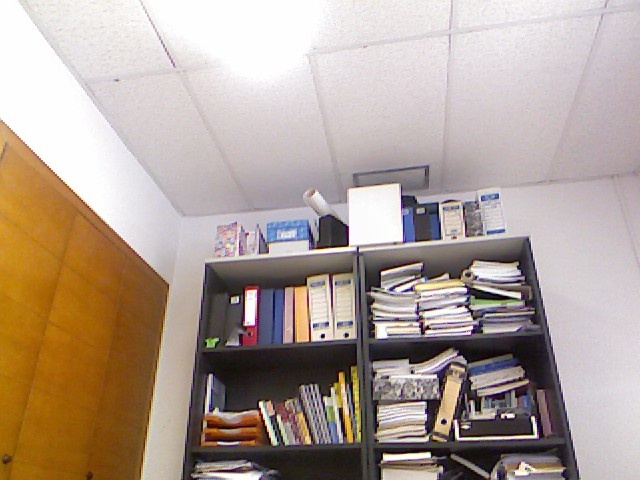} }
		\caption{\textbf{Lightswitch} sequence}
    \end{subfigure}\\
     \begin{subfigure}[b]{0.5\textwidth}
         {\includegraphics[width=1.1in]{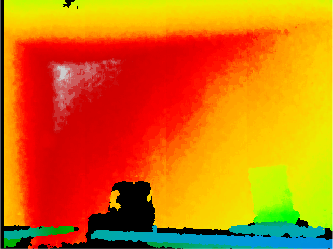}}
		 {\includegraphics[width=1.1in]{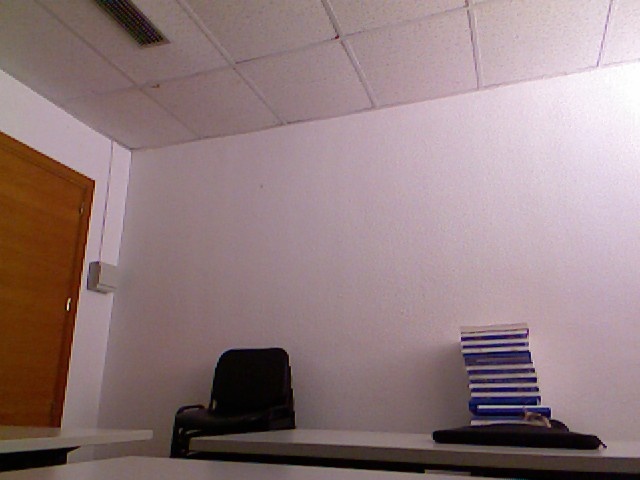} }
		{\includegraphics[width=1.1in]{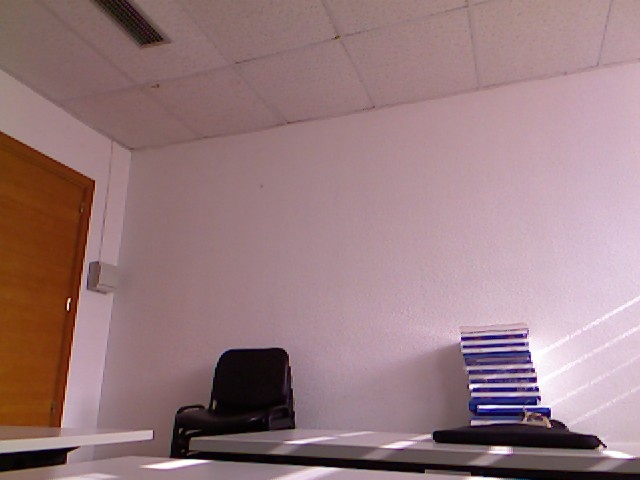} }
		\caption{\textbf{Time of day} sequence}
    \end{subfigure}\\
    \begin{subfigure}[b]{0.5\textwidth}
         {\includegraphics[width=1.1in]{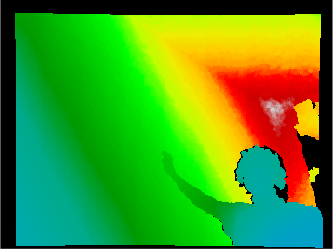}}
		 {\includegraphics[width=1.1in]{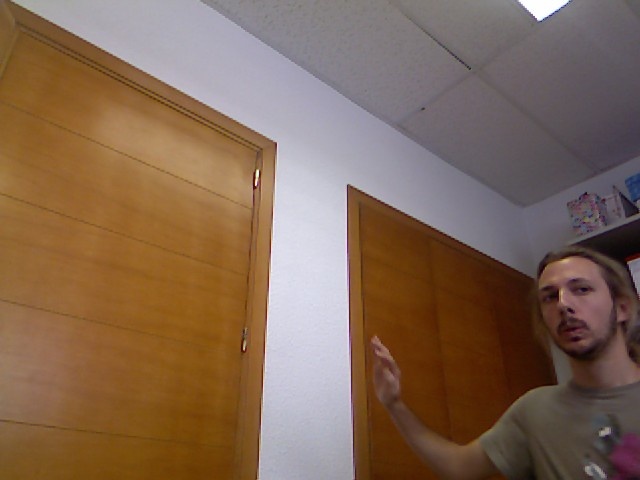} }
		{\includegraphics[width=1.1in]{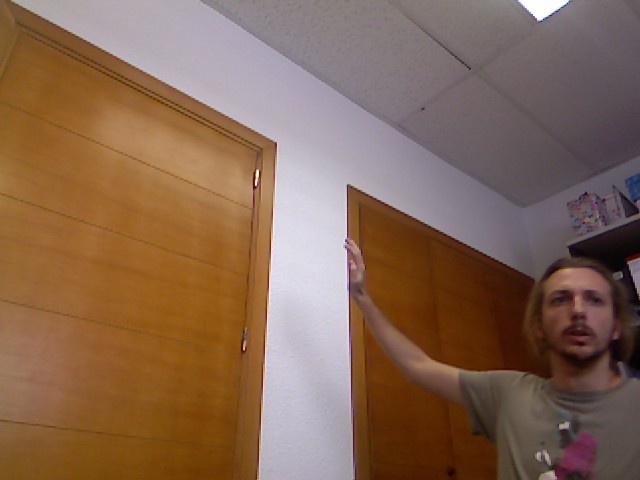} }
		\caption{\textbf{Shadow} sequence}
    \end{subfigure}\\
    \begin{subfigure}[b]{0.5\textwidth}
         {\includegraphics[width=1.1in]{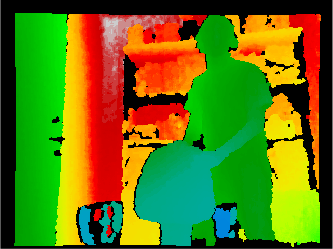}}
		 {\includegraphics[width=1.1in]{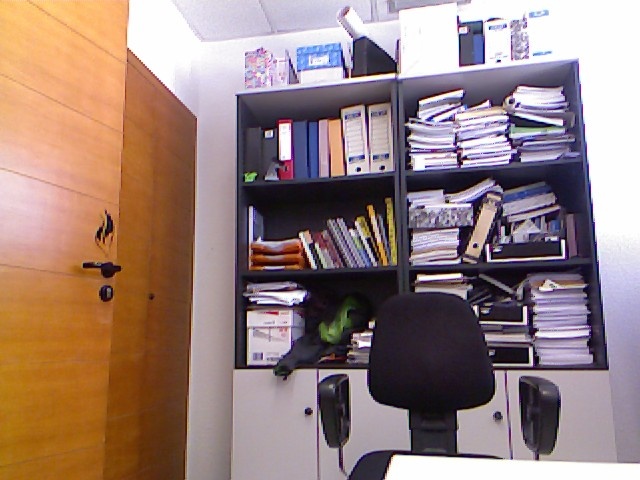} }
		{\includegraphics[width=1.1in]{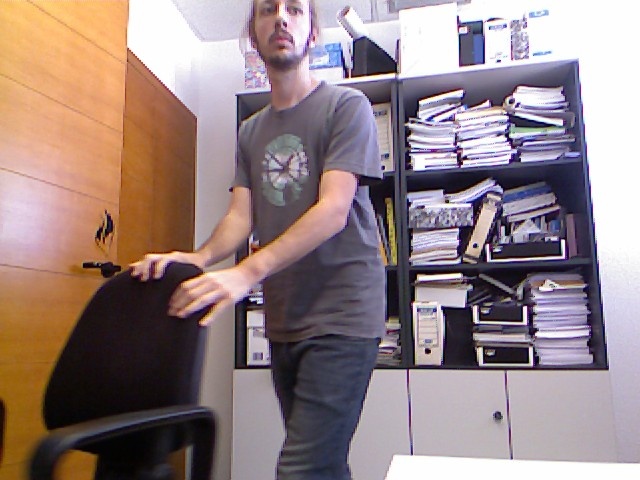} }
		\caption{\textbf{Waking object} sequence}
    \end{subfigure}\\
        \begin{subfigure}[b]{0.5\textwidth}
         {\includegraphics[width=1.1in]{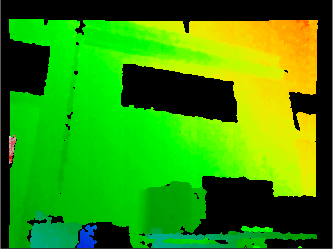}}
		 {\includegraphics[width=1.1in]{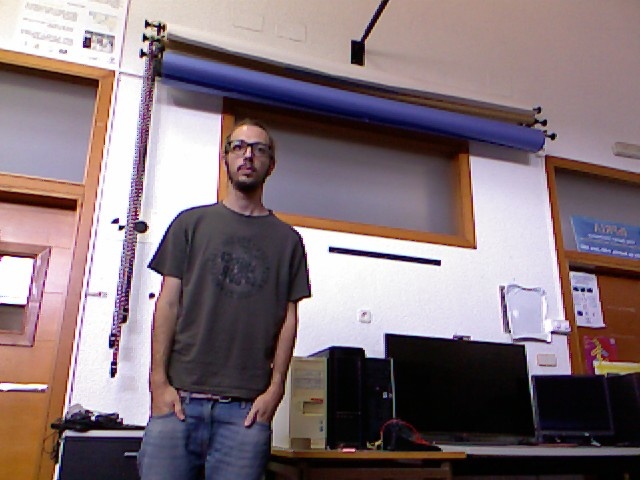} }
		{\includegraphics[width=1.1in]{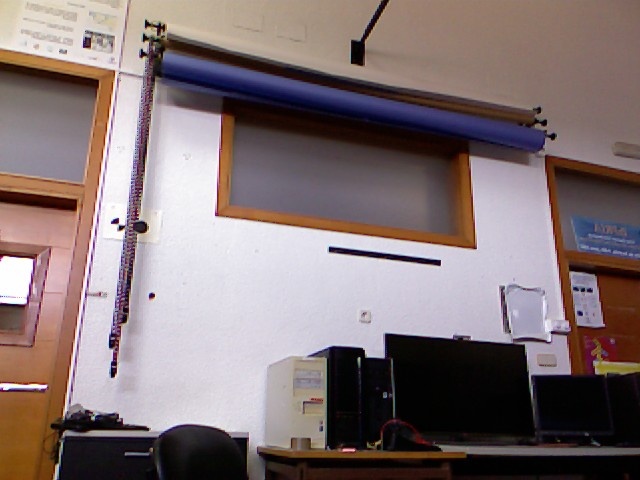} }
		\caption{\textbf{Bootstraping} sequence}
    \end{subfigure}\\
    \caption{Sequences of our new RGBD dataset. Each row shows the depth configuration of each scene and 2 different color frames.}
    	\label{fig_TestII}
\end{figure}

The Camplani dataset does not permit the proper evaluation of scene modeling algorithms due to the impossibility of evaluating over illumination changes, bootstrapping or waking objects, as there are no specific sequences with which to evaluate these issues. We built a new RGBD dataset to enable algorithm comparison and for generalization purposes. This dataset includes 7 different sequences (see Table~\ref{testII}) designed to test each of the main problems in scene modeling when both color and depth information are used. Each sequence starts with 100 training frames and has a hand-labelled foreground ground truth. In Fig.~\ref{fig_TestII}, examples of each sequence of the dataset can be found. Dataset and algorithm details can be found on \url{gsm.uib.es}.

\begin{table*} [!t]
\centering
\footnotesize
\caption{Characteristics of evaluated sequences from our dataset.}
\renewcommand{\arraystretch}{1.3}
\centering
\begin{tabular}{C{1.5cm}C{1.5cm}C{1.5cm}C{1.5cm}C{1.5cm}C{1.5cm}C{1.5cm}C{2.2cm}}
\hline 
\textbf{Sequence} & \textbf{Sleeping\_ds} & \textbf{TimeOfDay\_ds} & \textbf{Cespatx\_ds} & \textbf{Despatx\_ds} & \textbf{Shadows\_ds} &  \textbf{Ls\_ds} & \textbf{Bootstraping\_ds} \\ 
\hline 
\textbf{Number of frames} & 200 & 1231 & 428 & 465 & 330 & 407 & 300 \\ 
\hline 
\textbf{Number of ground truth frames} & 10 & 23 & 11 & 12 & 11 & 9 & 11\\ 
\hline 
\textbf{Test Objective} & Waking Object & Time of Day & Color camouflage & Depth camouflage & Shadows & Light Switch & Bootstraping\\ 
\hline 
\end{tabular} 
\label{testII}
\end{table*}

As for performance measures, we computed them using the framework of the CVPR 2014 \textit{CDnet} challenge~\cite{challengenet}, which implements the following seven different measures: recall, specificity, false positive ratio (FPR), false negative ratio (FNR), percentage of wrong classifications (PWC), precision and f-measure. Tables~\ref{UFresults} and~\ref{UBresults} show the specific results of GSM for each metric and sequence. 

A ranking of the tested algorithms is also computed, starting from the partial ranks on these measures (see Table~\ref{final_results}). We use the RM and the RC metrics, as in the previous evaluation. We evaluate our proposed algorithm against three different fusion algorithms: ViBe \cite{Barnich2011}, a mixture of Gaussians (MoG) implementation in the Opencv library by Zivkovic~\cite{zivkovic2004improved} and a background subtraction algorithm~\cite{Elgammal2000} that uses a Gaussian kernel (KDE). Following the \textit{CDnet} rules, each algorithm uses a single set of parameters.


It is important to notice that adding depth information leads to more robust scene modeling algorithms due the invariance of depth information to different types of illumination changes and the greater sensitivity of color information in cases of depth camouflage or ADO situations.

\begin{figure}[htpb]
\includegraphics[width=3.6in]{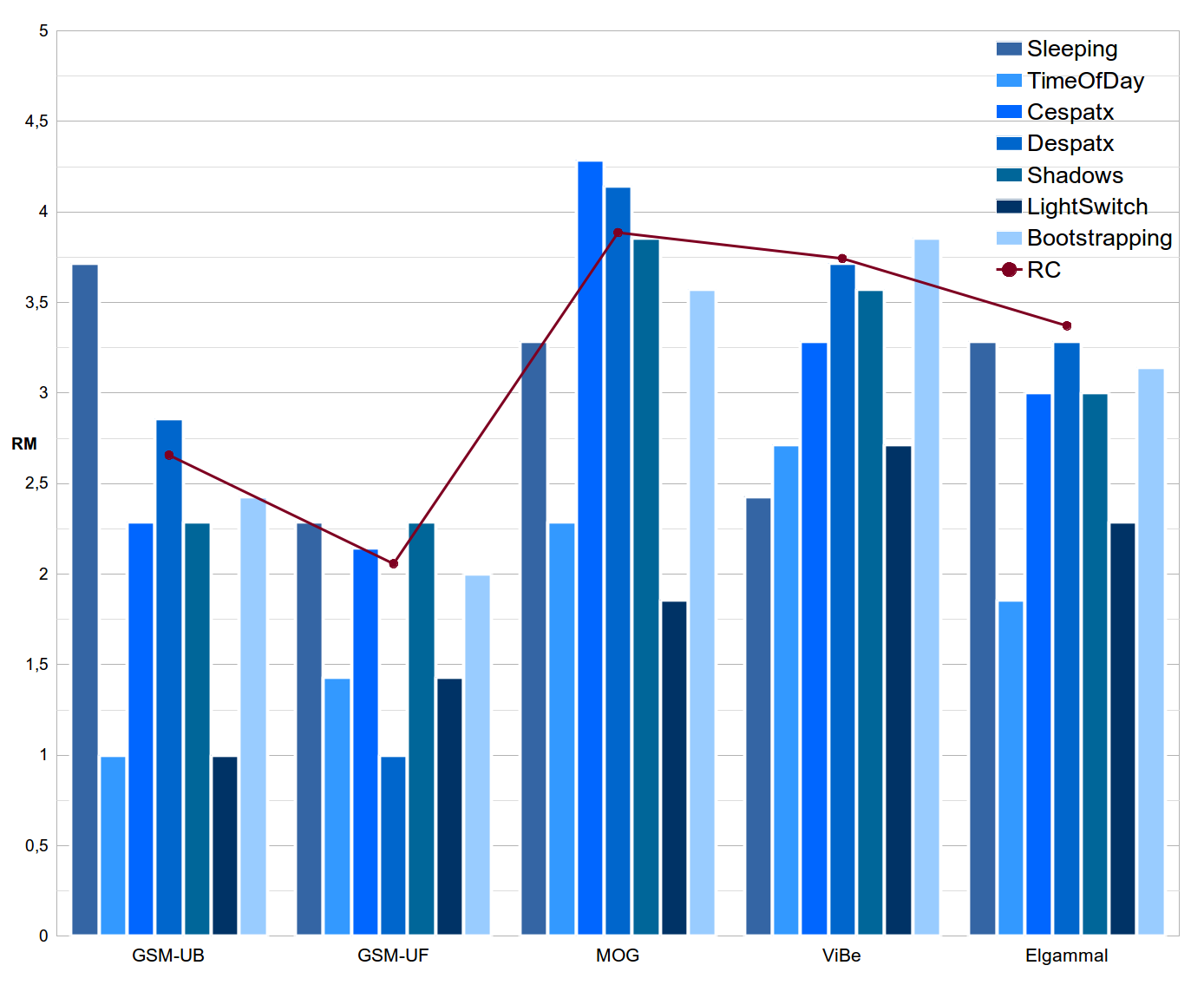}
\caption{GSM dataset simulation Results (\textbf{RM}). It can be found results of four different sequences for all tested algorithms and final comparisons. As 
GSM$_{{UF}}$ and  GSM$_{{UB}}$ are not the best for each sequence, are the most regular ones as it can be seen with the \textbf{RC} line, the lower the better.}
\label{fig_sim}
\end{figure}

To understand the global results, we analyze the performance of these algorithms for each sequence (see Fig~\ref{fig_sim}). The proposed algorithm has good results when we test different color situations. GSM$_{{UB}}$ and GSM$_{{UF}}$ prove to be the most stable under sudden illumination changes (\textit{Ls\_ds}), with significant difference from the other algorithms. In Depth camouflage (\textit{Despatx\_ds}) situations, it is important to notice that the results of KDE approach are very similar as to those of the proposed algorithm. This is because, in this sequence, the important information is the color information and we model that in the same way. 

The addition of one geometric dimension to our model permits us to obtain a small advantage in the evaluation of the Time of Day situations and in the Color Camouflage situations.

The Sleeping\_ds sequence  allows us to test whether the background object movement detection obtains the expected results. In \textit{Sleeping\_ds} sequence results from GSM$_{{UF}}$ and MOG are better than those from GSM$_{{UB}}$, as is shown in Table~\ref{final_results}. This occurs because in the last part of the sequence, the user is near the sensor provoking the apparition of a large region with ADO pixels. 

In the case of shadow evaluation (\textit{Shadows\_ds}), KDE algorithm has good results due to special treatment of color information to 
avoid shadows. GSM$_{{UF}}$ and GSM$_{{UB}}$ have the best results, proving that adding depth information can help to avoid some color 
problems (see Table~\ref{final_results}).

Our proposed algorithm has the best results in Bootstrapping situations (see Table~\ref{final_results}). This sequence is very challenging because, in the training stage, we have the assumption that depth information is constant over all frames; therefore, it is possible to model incorrect distributions, which leads to misclassification.

\begin{table*}[!t]
\footnotesize
\centering
\caption{Complete results for our proposed method, GSM$_{{UF}}$, for each category of the evaluation dataset.}

\begin{tabular}{ c c  c  c  c  c  c  c  }
\hline
	\textbf{GSM}$_{{UF}}$ & \textbf{Recall} & \textbf{Specificity}  & \textbf{FPR}  & \textbf{FNR}  & \textbf{PWC}  &\textbf{ F-Measure} & \textbf{Precision} \\ \hline\hline
	\textit{Sleeping} & 0.959 & 0.961 & 0.039 & 0.041 & 3.981 & 3.74 & 0.953 \\ 
	\textit{TimeOfDay} & 0 & 0.997 & 0.003 & 0 & 0.307 & 0 & 0 \\ 
	\textit{Color Camouflage} & 0.981 & 0.99 & 0.01 & 0.019 & 1.489 & 3.922 & 0.993 \\ 
	\textit{Depth Camouflage} & 0.971 & 0.989 & 0.011 & 0.029 & 2.008 & 3.878 & 0.988 \\ 
	\textit{Shadows} & 0.983 & 0.995 & 0.005 & 0.017 & 1.043 & 3.931 & 0.994 \\ 
	\textit{LightSwitch} & 0 & 0.997 & 0.003 & 0 & 0.343 & 0 & 0 \\ 
	\textit{BootStraping} & 0.85 & 0.995 & 0.005 & 0.15 & 3.907 & 3.493 & 0.979 \\ \hline
	\textbf{Average} & \textbf{0.630}  & \textbf{0.99}	& \textbf{0.01} & \textbf{0.08} & \textbf{3.67} & \textbf{2.58} & \textbf{0.71} \\ \hline

\end{tabular}
\label{UFresults}
\end{table*}

\begin{table*}[!t]
\footnotesize
\centering
\caption{Complete results for our proposed method, GSM$_{{UB}}$, for each category of the evaluation dataset.}

\begin{tabular}{ c c  c  c  c  c  c  c  }
\hline
	 \textbf{GSM}$_{{UB}}$& \textbf{Recall} & \textbf{Specificity}  & \textbf{FPR}  & \textbf{FNR}  & \textbf{PWC}  &\textbf{ F-Measure} & \textbf{Precision} \\ \hline\hline
	\textit{Sleeping} & 0.808 & 0.984 & 0.016 & 0.192 & 10.389 & 3.373 & 0.98   \\ 
	\textit{TimeOfDay} & 0 & 0.998 & 0.002 & 0 & 0.187 & 0 & 0  \\ 
	\textit{Color Camouflage} & 0.956 & 0.993 & 0.007 & 0.044 & 2.888 & 3.851 & 0.995   \\ 
	\textit{Depth Camouflage} & 0.941 & 0.992 & 0.008 & 0.059 & 3.388 & 3.796 & 0.991  \\ 
	\textit{Shadows} & 0.964 & 0.997 & 0.003 & 0.036 & 1.813 & 3.881 & 0.997 \\ 
	\textit{LightSwitch} & 0 & 0.999 & 0.001 & 0 & 0.114 & 0 & 0  \\ 
	\textit{BootStraping} & 0.743 & 0.996 & 0.004 & 0.257 & 6.941 & 3.19 & 0.984  \\ \hline
	\textbf{Average} &\textbf{ 0.68} & \textbf{0.99} & \textbf{0.01} & \textbf{0.04} & \textbf{1.58} & \textbf{2.71} & \textbf{0.70} \\ \hline

	 \end{tabular}
\label{UBresults}
\end{table*}

\begin{table*}[!htpb] 
\centering
\footnotesize
\caption{Evaluation results for all algorithms averaged over all sequences (\textbf{RM}). Last column shows final average ranking (\textbf{RC}). Bold entries indicate the 
best result and italics the second one.}
\begin{tabular}{  c  c  c  c  c  c  c  c  c }
\hline
Sequence	& \textbf{Sleeping\_ds} & \textbf{TimeOfDay\_ds} & \textbf{Cespatx\_ds}  & \textbf{Despatx\_ds} & \textbf{Shadows\_ds} & \textbf{Ls\_ds} & \textbf{Bootstraping\_ds} & \textbf{RC} \\
	&  &  & \textbf{Camouflage}    & \textbf{Camouflage} &  &  &  & 
 \  \\ \hline\hline
	\textbf{GSM-UB}  & 3             & \textbf{1}    & \textbf{2}     & \textit{2.857}& \textbf{2}& \textbf{1}    & \textit{2.429 }   & \textit{2.457} \\ 
	\textbf{GSM-UF}  & \textbf{1.857}& \textit{1.429}& \textit{2.714} & \textbf{2.714}& 2.571        & \textit{1.429 }&\textbf{2}& \textbf{2.429} \\ 
	\textit{MOG}~\cite{zivkovic2004improved}        & \textit{2.571}& 2.286         & 4.571          & 3.286         & 3.286         & 1.857         & 3.571    & 3.629 \\ 
	\textit{ViBe}~\cite{JeromeLeensSebastienPierardOlivierBarnich1978}     & 4.286         & 2.714         & \textit{2.714 }& 3.143         & 4.143         & 2.714         & 3.857    & 3.886 \\ 
	\textit{KDE}~\cite{elgammal2014background} & 3.286         & 1.857         & 3              & 3       & 3   & \textit{2.286} & 3.143    & 3.314 \\ \hline
\end{tabular}

\label{final_results}
\end{table*}

\section{Conclusion}
\label{sec:Conclusion}


We presented a new scene modeling approach, GSM, that uses both depth and color information in a unified way. We constructed a background model for each pixel of the scene and estimated the probability that a newly observed pixel value belongs to that model. These probabilities are estimated independently for each new frame.

We constructed our model using a Kernel Density Estimation (KDE) process with a Gaussian Kernel. To construct only one model, we used a three-dimensional kernel, with one dimension to model depth information and two for normalized chromaticity coordinates. We modelled sensor Absent Depth Observations (ADOs) using a probabilistic strategy to distinguish the pixels belonging to the background model from those which are provoked by foreground objects and detected each of these types of pixels. Pixels that cannot be classified as background or foreground were placed in a third classification class, which we called undefined, to classify these pixels.

We developed an algorithm to detect changing background objects in the same frame in which they move based on the cdf of the pixel model. Two updating strategies are used, to adapt the update phase to the different natures of the color and depth information. 

We provided all technical details to allow algorithm replication, including an algorithm description and the thresholds we used. We also constructed a new dataset (available at \url{gsm.uib.es}) to evaluate all background subtraction issues in related work, adding the new depth challenges.

Results show that the proposed algorithm is the most regular, having good results in a wide range of situations and solving the problems of the depth data sensors. This means that the algorithm can handle many different situations. We can conclude that the combination of two types of information in a 3D kernel helps to achieve better modeling algorithms.

The proposed algorithm has three different classes: background, foreground and undefined. To enable direct comparisons with the state-of-the-art algorithms, we decided to develop two implementations: {GSM}$_{{UF}}$ and {GSM}$_{{UB}}$.

Selection between the {GSM}$_{{UF}}$ and {GSM}$_{{UB}}$ implementations depends on the final application in which the scene modeling is used. Basically, we can distinguish two different situations: In applications where the changes occur at certain camera distances or when the scene tends to be static, as in surveillance applications, we recommend using the {GSM}$_{{UB}}$ implementation, as ADO tend to be provoked by remote parts of the scene, specular materials in the background and shadows that are reflected in the background of the scene. Instead, if the method is used in human interaction applications, such as tracking or human pose estimation, when the action occurs near the camera we recommend using the {GSM}$_{{UF}}$ implementation. In this case, ADO are normally provoked by near objects that appear during the sequence.


Our algorithm is designed following a per-pixel approach and is easily parallelizable because each pixel has its own model independent from the others. During the experimentation process, new RGBD sensors have appeared with more depth resolution. It could be interesting to test our algorithm with different devices. It is necessary to remark that our background subtraction algorithm is not designed only for with Microsoft Kinect. It can be adapted to different types information cues, such as thermal imagery.


%

%
%
%
\section*{Acknowledgment}
This work was partially funded by the Project TIN2012-35427 of the Spanish Government, with
FEDER support. 



%

\bibliographystyle{model2-names}
\bibliography{refs}

%

\end{document}